\documentclass[10pt,journal,compsoc]{IEEEtran}

\ifCLASSOPTIONcompsoc
  \usepackage[nocompress]{cite}
\else
  \usepackage{cite}
\fi

\ifCLASSINFOpdf
  \usepackage[pdftex]{graphicx}
\else
\fi

\usepackage{grffile}

\usepackage{amsmath,amsfonts}

\usepackage{algorithmic}
\usepackage{algorithm}

\usepackage{array}

\ifCLASSOPTIONcompsoc
 \usepackage[caption=false,font=footnotesize,labelfont=sf,textfont=sf]{subfig}
\else
 \usepackage[caption=false,font=footnotesize]{subfig}
\fi

\usepackage{url}

\usepackage[capitalize,noabbrev]{cleveref}
\usepackage{xcolor}
\usepackage{xspace}
\usepackage{multirow}
\usepackage{colortbl}
\usepackage{wrapfig}
\usepackage{bm}
\usepackage{booktabs}
\newcommand{\softmax}{\mathrm{softmax}}
\newcommand{\myparagraph}[1]{\vspace{2pt}\noindent{\bf #1}}
\newcommand{\elodi}{\textsc{Elodi}\xspace}
\newcommand{\eg}{\emph{e.g}\xspace}
\newcommand{\etal}{\emph{et al.}\xspace}
\newcommand{\ie}{\emph{i.e}\xspace}
\newcommand{\iid}{\emph{i.i.d}\xspace}
\newcommand{\wrt}{\emph{w.r.t}\xspace}
\newcommand{\aka}{\emph{a.k.a}\xspace}
\newcommand{\versus}{\emph{v.s}\xspace}

\def\loss{\mathcal{L}}
\def\model{\mathcal{M}}
\mathchardef\mhyphen="2D
\newcommand{\unsim}{\mathord{\sim}}  %

\usepackage{tikz}
\usetikzlibrary{cd}

\usepackage[T1]{fontenc}%
\usepackage{xhfill}%
\newcommand{\ditto}[1][.4pt]{\xrfill{#1}~\textquotedbl~\xrfill{#1}}

\usepackage{pifont}
\newcommand{\smark}{\ding{110}\xspace}%
\newcommand{\trimark}{\ding{115}\xspace}%

\usepackage{soul}
\sethlcolor{Gray}

\def\vphi{\bm{\phi}}
\def\vpsi{\bm{\psi}}
\def\vzero{{\bm{0}}}
\def\vmu{{\bm{\mu}}}
\def\mSigma{{\bm{\Sigma}}}
\def\vs{{\mathbf{s}}}
\def\vz{{\mathbf{z}}}
\def\cF{{\mathcal{F}}}
\def\cH{{\mathcal{H}}}
\def\sK{{\mathbb{K}}}
\def\sM{{\mathbb{M}}}
\def\sX{{\mathbb{X}}}
\def\Normal{\mathcal{N}}

\usepackage{wasysym}
\newcommand{\half}{$^{\tiny\RIGHTcircle}$} %
\newcommand{\full}{$^{\tiny\Circle}$} %
\newcommand{\distill}{$^\diamondsuit$}
\newcommand{\lowbetter}{$_\downarrow$}

\definecolor{Gray}{gray}{0.9}

\definecolor{Darkgray}{rgb}{0,0,0}
\definecolor{tabgreen}{rgb}{0,0,0}
\definecolor{tabred}{rgb}{0,0,0}
\definecolor{taborange}{rgb}{0,0,0}
\definecolor{tabpurple}{rgb}{0,0,0}
\definecolor{red}{rgb}{0,0,0}
\definecolor{brown}{rgb}{0,0,0}
\definecolor{blue}{rgb}{0,0,0}
\definecolor{magenta}{rgb}{0,0,0}

\begin{document}
\title{\elodi: Ensemble Logit Difference Inhibition for Positive-Congruent Training}

\author{Yue~Zhao$^*$,
        Yantao~Shen$^\dagger$,
        Yuanjun~Xiong,
        Shuo~Yang,
        Wei~Xia,
        Zhuowen~Tu,~\IEEEmembership{Fellow,~IEEE,}
        Bernt~Schiele,~\IEEEmembership{Fellow,~IEEE,}
        and~Stefano~Soatto,~\IEEEmembership{Fellow,~IEEE,} %
\IEEEcompsocitemizethanks{\IEEEcompsocthanksitem Y. Zhao is with Department
of Computer Science, University of Texas at Austin, Texas, 78712. $^*$The work was done during his internship at AWS AI Labs. 
E-mail: yzhao@cs.utexas.edu \protect\\
\IEEEcompsocthanksitem Y. Shen, Y. Xiong, S. Yang, W. Xia, Z. Tu, B. Schiele, and S. Soatto are with AWS AI Labs. $^\dagger$Corresponding author.
E-mail: yantaos@amazon.com
\protect\\
}%
\thanks{Manuscript received April 19, 2005; revised August 26, 2015.}}

\markboth{Journal of \LaTeX\ Class Files,~Vol.~14, No.~8, August~2015}%
{Shell \MakeLowercase{\textit{et al.}}: Bare Demo of IEEEtran.cls for Computer Society Journals}

\IEEEtitleabstractindextext{%
\begin{abstract}
Negative flips are errors introduced in a classification system when a legacy model is updated. Existing methods to reduce the negative flip rate (NFR) either do so at the expense of overall accuracy by forcing a new model to imitate the old models, or use ensembles, which multiply inference cost prohibitively. We analyze the role of ensembles in reducing NFR and observe that they remove negative flips that are typically not close to the decision boundary, but often exhibit large deviations in the distance among their logits. Based on the observation, we present a method, called Ensemble Logit Difference Inhibition (\elodi), to train a classification system that achieves paragon performance in both error rate and NFR, at the inference cost of a single model. The method distills a homogeneous ensemble to a single student model which is used to update the classification system.
\elodi also introduces a generalized distillation objective, Logit Difference Inhibition (LDI), which only penalizes the logit difference of a subset of classes with the highest logit values.  
On multiple image classification benchmarks, model updates with \elodi demonstrate superior accuracy retention and NFR reduction. 
\end{abstract}

}

\maketitle

\IEEEdisplaynontitleabstractindextext

\IEEEpeerreviewmaketitle

\IEEEraisesectionheading{\section{Introduction}\label{sec:introduction}}

\IEEEPARstart{T}{he} rapid development of visual recognition in recent years has led to the need for frequently updating existing models in production-scale systems. However, when replacing a legacy classification model, one has to weigh the benefit of decreased error rate against the risk of introducing new errors that may disrupt post-processing pipelines \cite{yan2021pct} or cause friction with human users \cite{bansal2019updates}.
To address this, we study Positive-Congruent Training (PC-Training) which refers to any training procedure that minimizes the {\em negative flip rate} (NFR) along with the error rate (ER) simultaneously. 

Negative flips are instances that are misclassified by the new model but correctly classified by the old one. They are manifest in both visual and natural language tasks~\cite{yan2021pct,xie2021regression}. They typically include {\em not only} samples close to the decision boundary, but also high-confidence mistakes that lead to  {\em perceived ``regression''} in performance compared to the old model.
They are present even in identical architectures trained from different initial conditions, with different data augmentations, or using different sampling of mini-batches.
Yan~\etal~\cite{yan2021pct} have shown that in state-of-the-art image classification models, where a 1\% improvement is considered significant, NFR can be in the order of 4$\unsim$5\% even across models that have identical ER.
These intriguing properties motivate us to investigate the causes of negative flips and the mechanism of reducing negative flips to establish a model update method that achieves cross-model compatibility, thus lowering NFR and ER, for better positive-congruent training.

\myparagraph{Two questions.}
A naive approach to cross-model compatibility is to bias one model to mimic the other, as done in model distillation~\cite{hinton2015kd}. In this case, however, compatibility comes at the expense of accuracy~\cite{yan2021pct,bansal2019updates}. On the other hand,  averaging a number of models in a deep ensemble~\cite{lakshminarayanan2017deepensemble} can reduce NFR without negative accuracy impact~\cite{yan2021pct}, even if it does not explicitly optimize NFR nor its surrogates. The role of ensembles in improving accuracy is widely known, but our first question arises: {\em what is the role of ensembles in reducing NFR}?

Even though using deep ensembles achieves state-of-the-art performance in terms of reducing NFR~\cite{yan2021pct}, it is nonviable in real applications at scale since it multiplies the cost of inference by an integer factor. Therefore, a second key question arises: {\em Is it possible to achieve the PC-Training performance of ensembles at the inference cost of a single model?}

\begin{figure*}[!t]
	\centering
	\subfloat[]{
		\includegraphics[width=0.45\linewidth]{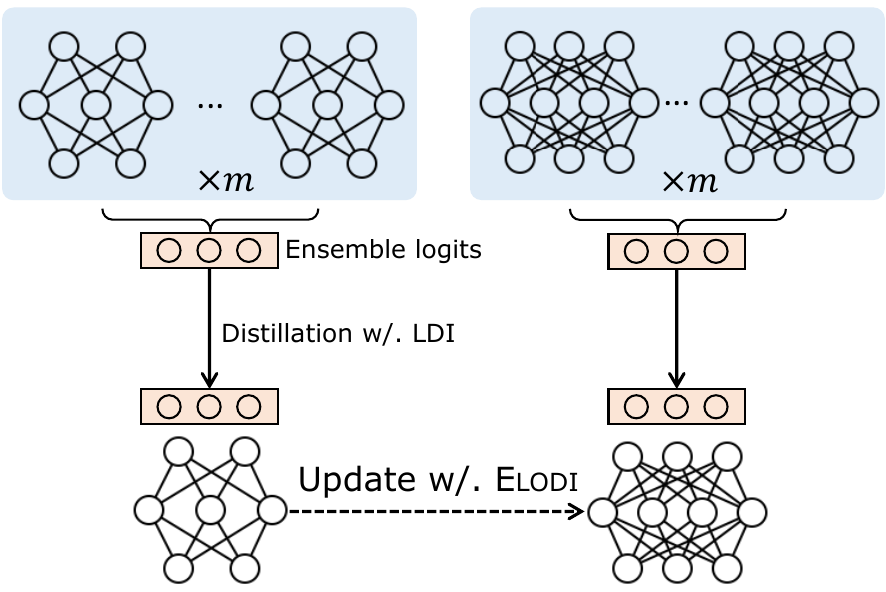}%
		\label{fig:teaser:a}
	}
	\hfil
	\subfloat[]{
		\includegraphics[width=0.52\linewidth]{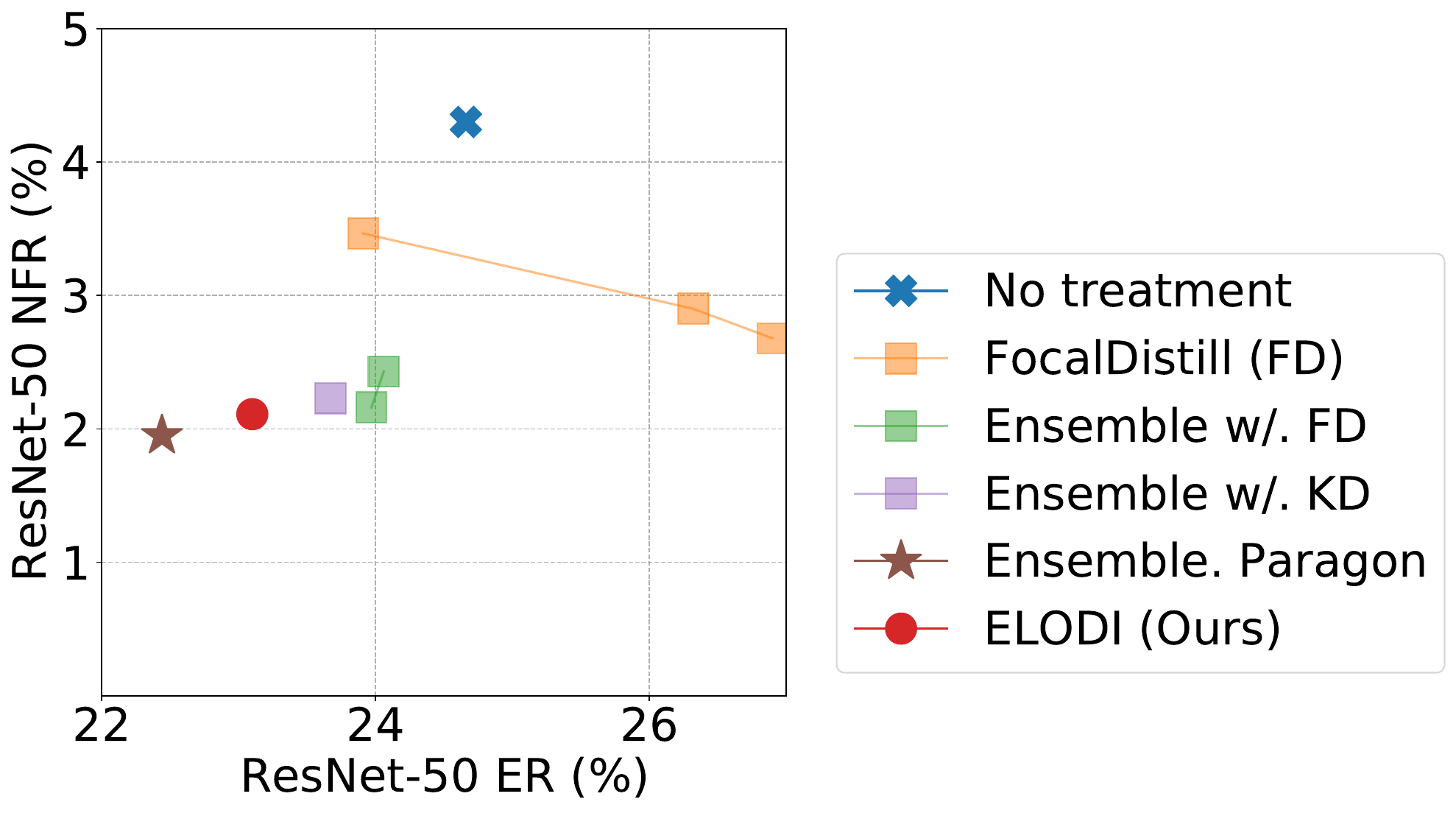}%
	}
	\caption{
        \textbf{The overview of the proposed Ensemble Logit Difference Inhibition (\elodi) method and its effectiveness compared to baseline methods.}
		When updating an old model to a new one, we aim to minimize the negative flip rate (NFR), namely the ratio of instances that are misclassified by the new model but correctly classified by the old one over all samples, and the error rate (ER) simultaneously.
        \textbf{(a)} In \elodi, for both the old and new models that we wish to deploy, we first train $m$ replicas of the same architectures on the same training data with different network initializations to form a deep ensemble.
        Next, we train both the old and new models to deploy using the Logit Difference Inhibition (LDI) loss with respect to the ensemble of $m$ models of their same architecture.
		The result is a new model that achieves a significantly reduced NFR compared to the old model.
		\textbf{(b)} We look at an example where the old model is a ResNet-18 and the new model is a ResNet-50 and present the scatter plot of the ResNet-50's ER and its NFR \wrt the ResNet-18. 
		The more left and lower, the better.
		\elodi improves both ER and NFR than baseline methods.
        The numbers can be found in~\cref{tab:main_result}.
        Particularly, {\textcolor{red}{\elodi}} is close to the \textcolor{brown}{ensemble paragon}, without the prohibitive computation cost of ensembles. 
	}
	\label{fig:teaser}
\end{figure*}

\myparagraph{Key ideas.}
To address the {\em first} key question above, we analyze the pattern of negative flip reduction in deep ensembles. We observe that deep ensembles reduce NFR by remedying potential flip samples that have relatively large variations in the logits space of different single models. When a deep ensemble is composed of member models with the same architecture but trained with independent initialization on the same dataset, which we denote as {\em homogeneous ensembles}, this behavior can be theoretically predicted and empirically validated.

As illustrated in \cref{fig:teaser:a},
we independently train replicas of a single model with different random seeds to form the deep ensemble. We introduce a generalized distillation objective, Logit Difference Inhibition (LDI), which only penalizes the logit difference between the reference ensemble and the student single model on a subset of classes with the highest logit values.
The result is what we call {\em Ensemble Logit Difference Inhibition} (\textbf{\elodi}).

\myparagraph{Contributions.} \elodi improves the state of the art in reducing perceived regression in model updates in three ways: (1) Generality, by not targeting distillation to a specific
legacy model, yet reducing NFR; (2) Absence of collateral damage, by retaining the accuracy of a new model, or even improving it, while ensuring reduction of
NFR; (3) Efficiency, as \elodi does not require evaluating ensembles of models at inference time.
Moreover, \elodi is compatible with existing models trained without treatment.
These improvements are made possible by two main contributions: (1) an analysis on deep ensembles which sheds light on their role in reducing NFR and the direction to obtain their performance for PC-training with single models; (2) \elodi, that integrates the NFR reduction of deep ensembles and running cost of single models by first training deep networks using the LDI loss with respect to an ensemble and then deploying the resulting single model at inference time. This results in a significant reduction of NFR (29\% relative reduction on ImageNet for ResNet-18 $\rightarrow$ ResNet-50) over previous methods. As a side benefit, \elodi increases top-1 accuracy in several cases, and is comparable in others.
Code is publicly available at \url{https://github.com/amazon-science/regression-constraint-model-upgrade} to facilitate future research in this emerging field.

\section{Related Work}
\label{sec:related}
\myparagraph{Cross-model compatibility} is becoming increasingly important as real-world systems incorporate trained components that, if replaced, can wreak havoc with post-processing pipelines. Toneva~\etal~\cite{toneva2019empirical} empirically study prediction flip on training samples between epochs, termed ``forgetting events'', while Yan~\etal\cite{yan2021pct} address perceived regression using held-out sets between different models. Both are particular instances of cross-model compatibility~\cite{shen2020bct,bansal2019updates,srivastava2020empirical}.
Focal Distillation~\cite{yan2021pct} minimizes the distance between the old and new predictions, with increased weights on samples correctly classified by the old model.
Jiang~\etal conduct experiments on more architectures and modalities including image, text, and tabular data in~\cite{jiang2022churn}.
\cite{trauble2021backward} use a probabilistic approach to determine whether the prediction should update when a new model comes.
While it improves {\em cumulative} NFR, it requires multiple models to be available at inference, which is prohibitive in practice.

\myparagraph{Ensemble learning} methods \cite{breiman1996bagging,freund1997adaboost,breiman2001randomforest} are widely adopted in machine learning.
The understanding of these methods is sometimes explained as enlarging the margins \cite{bartlett1998boosting}.
Recently, the ``multi-view'' hypothesis~\cite{allen2023towards} suggests that each independent model in an ensemble of deep networks learns a subset of feature views and memorizes data not separable using this subset.
In practice, one can always boost the performance of a classifier by averaging multiple models that are trained separately under a certain level of variation in training including model type, training data, initialization, etc.
In this paper, we take a different aspect of ensembling to reduce NFR, not to improve accuracy.
We apply the ensemble as a teacher's model to guide the student model in reducing negative flips during model updates.
In particular, we present an alternative explanation from the perspective of representations' dispersion in the logit space.
\elodi can be thought of as variance reduction regularization in a Bayesian NN ensemble, which is replaced by its mean at inference time.
The literature on variance reduction is too vast to survey here, but relevant references include \cite{hoeting1999bayesian,fragoso2018bayesian}.

Some other ensemble learning techniques are summarized as follows: Deep ensemble~\cite{lakshminarayanan2017deepensemble} improves accuracy and allows estimating sample uncertainty; Snapshot Ensemble~\cite{huang2017snapshot} and Fast Geometric Ensemble~\cite{garipov2018loss} train compo-
nent models simultaneously, and Yan~\etal~\cite{yan2021pct} show that ensembles help reduce regression. Ensembles are impractical in most real applications due to the multiplier they impose on inference costs. This has prompted research on ``implicit ensembles'' such as Dropout~\cite{srivastava2014dropout} and its variants~\cite{gal2016dropout}, DropPath~\cite{larsson2017fractalnet} and Stochastic Depth~\cite{huang2016deep}.
Wen~\etal propose BatchEnsemble~\cite{wen2020batchensemble} to generate ensemble weights, Havasi~\etal use a MIMO~\cite{havasi2021mimo} design to train multiple sub-
networks concurrently. Different from all these methods, our method mainly focuses on reducing NFR instead of improving accuracy.

\myparagraph{Knowledge distillation} (KD)~\cite{hinton2015kd} was proposed to transfer ``dark'' knowledge from a larger ``teacher'' network  to a smaller ``student'' by minimizing the distance between the distribution of predictions.
In self-distillation~\cite{zhang2019selfdistill}, the teacher and student are the same.
Focal Distillation~\cite{yan2021pct} is a special case of KD with a sample-specific filtering function, developed for model updates where the legacy ``teacher'' model is actually weaker than the student (new) model, as in {\em Reversed KD}~\cite{yuan2020revisiting}, where it is used as regularization.
Ensemble distillation uses multiple teachers to improve accuracy in vision and other applications~\cite{reich2020ensemble,fukuda2017efficient,asif2020ensemble,malinin2020ensemble,lin2020ensemble}.
Our method is related to ensemble distillation while having two distinctive differences: (1) Our method uses a different term for the loss to achieve reduction of NFR; (2) ours uses a \emph{homogeneous ensemble} whose members have the same architecture and are trained on the same dataset with different initialization seeds, unlike the traditional case that uses diverse models in the ensemble~\cite{kuncheva2003measures}, which we call a {\em heterogeneous ensemble}.

\section{Representation Landscape of Ensemble-based PC-Training}
\label{sec:probe}

\noindent To answer the first key questions in \cref{sec:introduction}, we explore (1) how negative flips occur
and (2) why ensembles yield fewer negative flips.
To do so, we analyze the so-called {\em logit space}, where the representations are computed by a deep network before the $\softmax$ operation live. 
The reason we analyze logits rather than feature or softmax probabilities is as follows.
Compared to the feature space,
the logits of an arbitrary sample produced by different models trained on the same dataset live in the same vector space which is defined by the label set of the training samples.
Compared to the space of post-$\softmax$ output, the logit distribution is easier to analyze because the $\softmax$ operation will skew the distribution.
From a practical perspective, averaging in the logit space for ensembles is also common in recent works~\cite{gontijo2022no,wortsman2022modelsoups}.

\subsection{Negative Flips and Logit Displacement Magnitude}
\label{sec:probe:landscape}

\myparagraph{Negative Flips}.
Given an input image $x$ with its label $\ell$ and a learned model $\vphi$, let $\vphi(x)\in\mathbb{R}^{C}$ denote its output logit vector before $\softmax$, and $\vphi_{k}(x)$ denote the $k$-th element, where $k\in\{1, \cdots, C\}$ and $ C $ is the number of classes.
The logit vector varies across models due to architecture, initialization, optimization method, and training dataset to name a few.
For any model pair $\vphi$ and $\vpsi$, where $\vphi$ and $\vpsi$ take different forms due to architectural change in general, we define the \emph{logit displacement} to be the difference between two output logits,~\ie$\vphi(x) - \vpsi(x)$.
Once the displacement is large enough to change the order of the top predictions, namely $\arg\max_k\vphi_k(x) \neq \arg\max_k\vpsi_k(x)$, a flip occurs.
We are particularly interested in {\em negative flips}, where we assume $\vphi$ to be an old model and $\vpsi$ to be a new one and 
$\arg\max_k\vphi_k(x)=\ell$ while $\arg\max_k\vpsi_k(x)\neq\ell$.

\myparagraph{Homogeneous ensembles.}
Let $\sM_1 = \{\cdots, \vphi^{(i)},\cdots\} $ and $\sM_2=\{\cdots, \vpsi^{(j)},\cdots\}$ denote the set of member models from two homogeneous ensembles:
Each $\vphi^{(i)}$ has the same model architecture and is trained on the same dataset despite being independently initialized.
So does each ${\vpsi}^{(i)}$, though $\vphi$ and $\vpsi$ can have different architectures
For simplicity, we consider $\|\sM_1\|=\|\sM_2\|=m$, where $\| \cdot \| $ is the cardinality of each set.

\myparagraph{Homogeneous ensembles reduce negative flips by reducing the magnitude of logit displacement.}
Given an input image $x$, $\{\vphi^{(1)}(x),\cdots,\vphi^{(m)}(x)\}$ can be considered as $m$ \iid random variables drawn from a distribution approximated to second-order by an expectation $\vmu_1$ and a co-variance matrix $\mSigma_1$,~\ie $\vphi(x)\sim \mathcal{D}(\vmu_1, \mSigma_1)$.
Likewise we also have $\vpsi(x)\sim \mathcal{D}(\vmu_2, \mSigma_2)$.
The ensembles' logit vectors are computed by averaging the individual models' logits and denoted as $\vphi^{(\mathrm{ens})}(x)$ and $\vpsi^{(\mathrm{ens})}(x)$.

The multi-dimensional central limit theorem~\cite{rvavceva1962domains,ferguson2017course} states that this average converges in distribution to a multivariate normal distribution with the increase of $m$,~\ie
\begin{align}
\label{eq:clt}
\vphi^{(\mathrm{ens})}(x) = \frac{1}{m}\sum_{i=1}^{m}\vphi^{(i)}(x) \overset{D}{\sim}\Normal(\vmu,\frac{1}{m}\mSigma).
\end{align}

Therefore, the logit displacement between the two ensembles converges in distribution to another multivariate normal distribution,~\ie
\begin{align}
\label{eq:clt_different}
\vphi^{(\mathrm{ens})}(x) - \vpsi^{(\mathrm{ens})}(x)
&= \frac{1}{m}\sum_{i\in\sM_1}\vphi^{(i)}(x) - \frac{1}{m}\sum_{j\in\sM_2}\vpsi^{(j)}(x) \\
& \overset{D}{\sim}\mathcal{N}\left(\vmu_1 - \vmu_2,\frac{\mSigma_1+\mSigma_2}{m}\right).
\end{align}
The norm of logit displacement will follow a generalized $\chi^2$ distribution~\cite{mathai1992quadratic,das2021method}.

As a special case, if $\sM_1$ and $\sM_2$ have the same model architecture, then we have a normal distribution with \emph{zero} mean and co-variance inversely scaled by the ensemble size:
\begin{align}
\label{eq:clt_same}
\vphi^{(\mathrm{ens})}(x) - \vpsi^{(\mathrm{ens})}(x)\overset{D}{\sim}\mathcal{N}\left(\vzero, \frac{\mSigma_1+\mSigma_2}{m}\right).
\end{align}

\begin{figure}[!tb]
    \centering
    \noindent\includegraphics[width=0.7\linewidth]{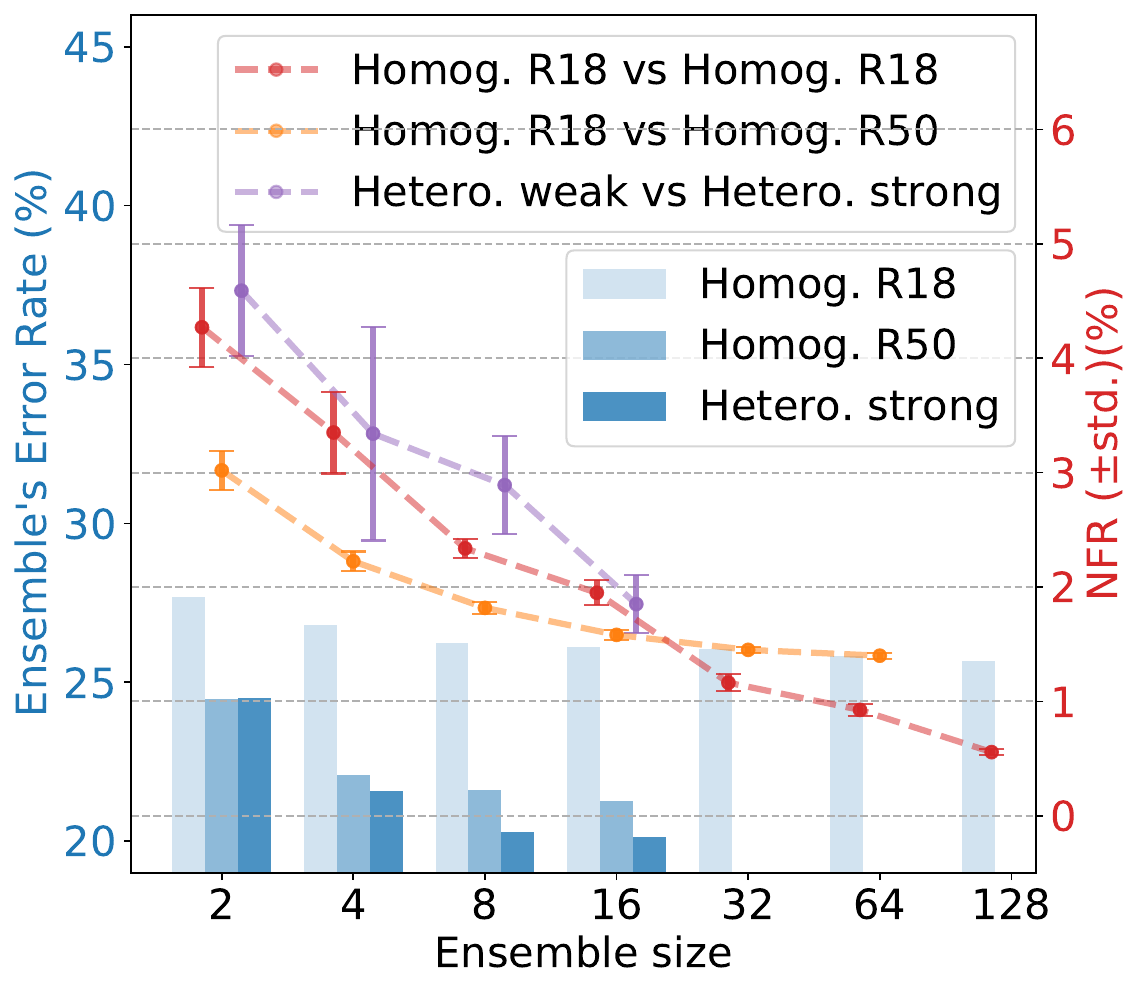}
    \caption{
            \textbf{ER/NFR \wrt ensemble size}.
            Our hypothesis explains the behaviors of model updates under different ensemble settings.
            Please refer to the second last paragraph in \cref{sec:probe:landscape}.
    }%
    \label{fig:homo_vs_hetero}
\end{figure}

We connect the analysis to the observations shown in~\cref{fig:homo_vs_hetero}:
(1) \cref{eq:clt_same} implies that when ensembles become larger, the expectation of logit difference is zero and the covariance keeps decreasing, resulting in consistently decreasing NFR.
This is consistent with the observation in \cite{yan2021pct}, which is redrawn in the {\color{tabred} red curve}, that two very large ensembles with the same architecture can have almost no flips.
(2) In the case of two homogeneous ensembles with different architectures, $\vmu_1$ and $\vmu_2$ could have a non-zero difference which results in NFR not converging to zero. But the decrease of covariance part in \cref{eq:clt_different} still contributes to a consistent non-trivial reduction in NFR. These explain the observation in the {\color{taborange} orange curve} that NFR stagnates at a smaller non-zero value.

\myparagraph{Heterogeneous ensembles show significantly higher variance on negative flips.}
Additionally, although we cannot conduct a similar analysis for a heterogeneous ensemble, we can empirically verify that the NFR between two heterogeneous ensembles has a significantly higher variance, as shown in the {\color{tabpurple} purple curve}, which suggests they may be inferior for NFR reduction compared to homogeneous ones.

\begin{figure*}[t]
	\begin{center}
		\subfloat[
        ]{
		\includegraphics[width=0.22\linewidth]{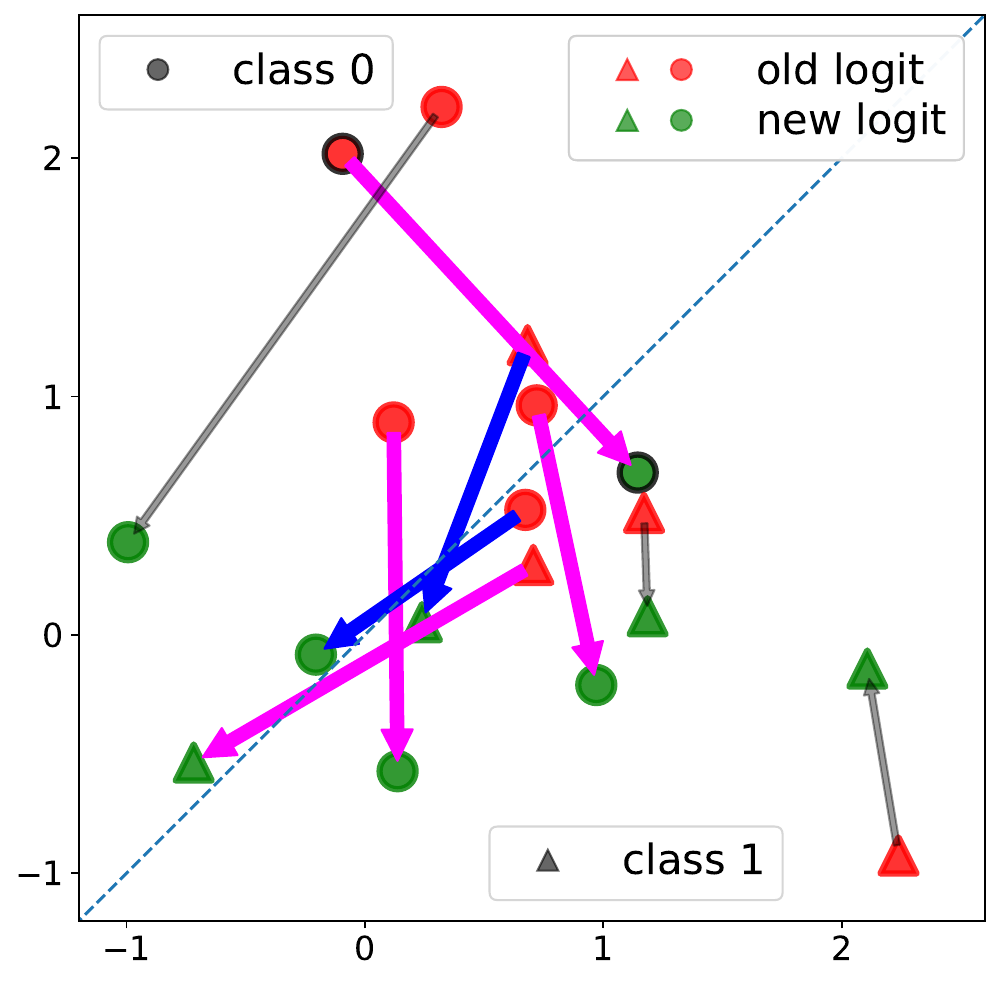}
		\label{fig:toy_model_single}
	}
		\hskip 0.08 in
		\subfloat[
        ]{
		\includegraphics[width=0.22\linewidth]{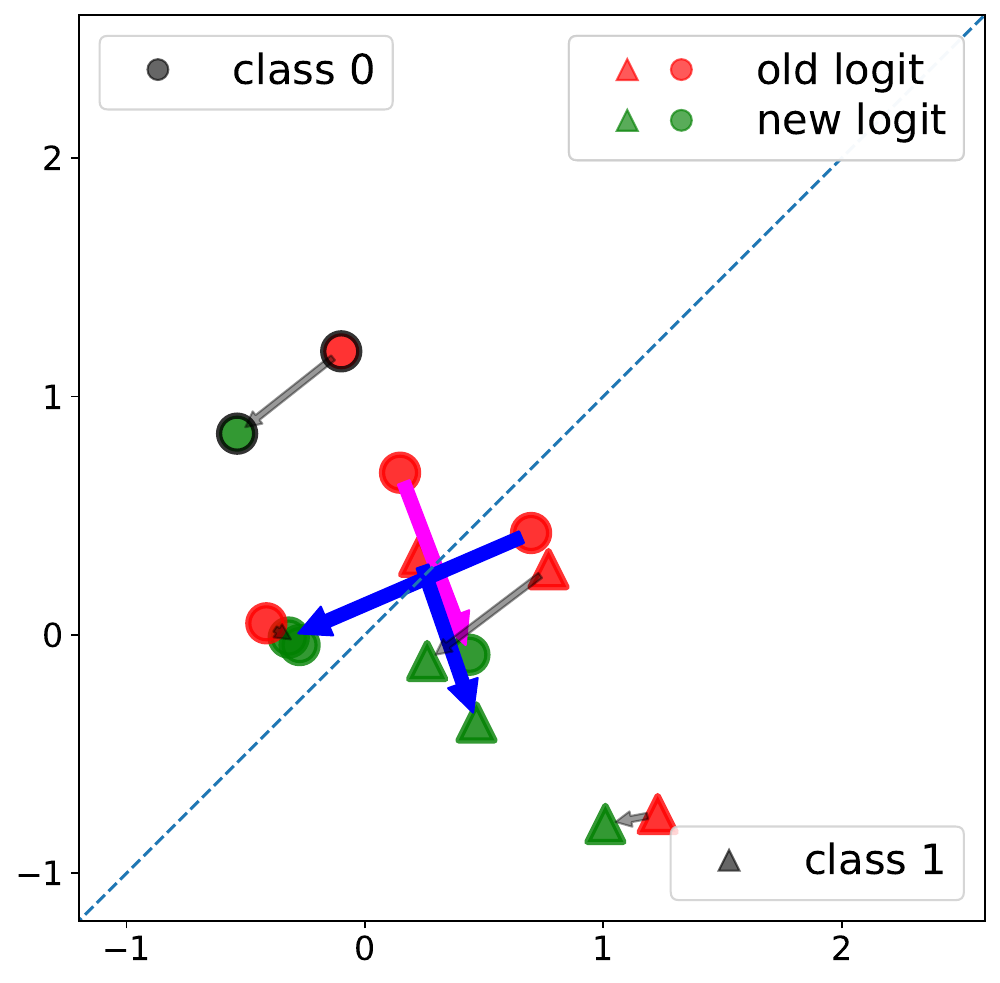}
		\label{fig:toy_model_ensemble}
	}
		\hskip 0.08 in
		\subfloat[
        ]{
		\includegraphics[width=0.22\linewidth]{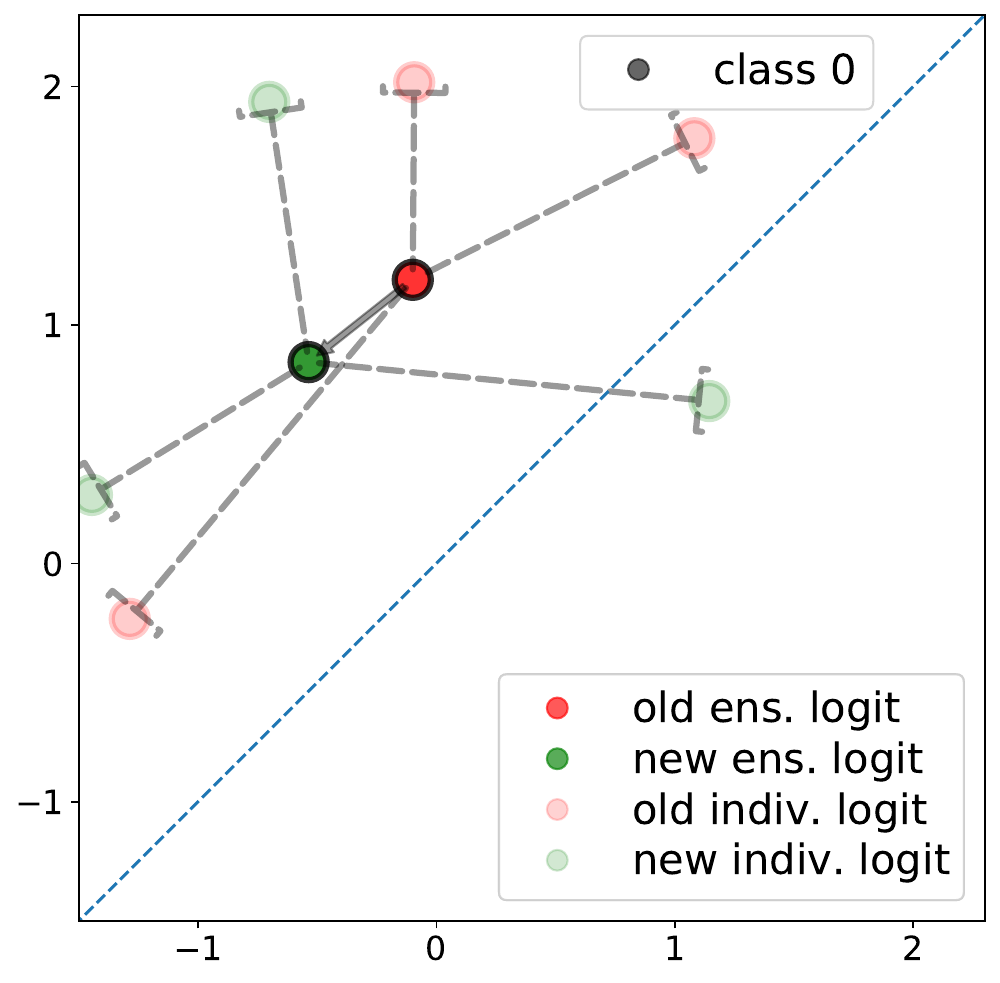}
		\label{fig:toy_model_drift}
	}
		\hskip 0.08 in
		\subfloat[
        ]{
        \includegraphics[width=0.21\linewidth]{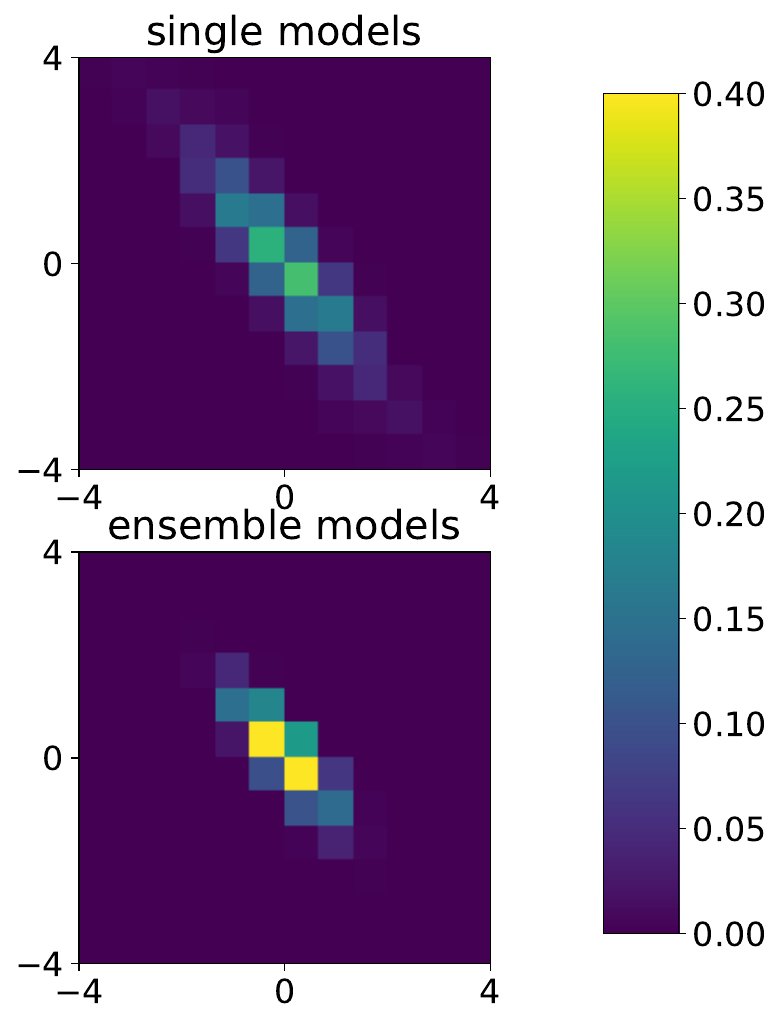}
        \label{fig:hist2d_clt}
	}
		\caption{
			\textbf{Visualization of a 2-class example}.
			\textbf{(a-c)} Two-class logits of two single models and/or ensembles. 
			\trimark and  $\bullet$ refer to the ground-truth classes, while
			{\textcolor{tabred}{red}} and {\textcolor{tabgreen}{green}} data points refer to old and new model's logits.
			{\textcolor{magenta}{Magenta arrow}}, {\textcolor{blue}{blue arrow}}, and {\textcolor{Darkgray}{gray arrow}}
			link negative flip, positive flip, and consistent (either both correct or both wrong) prediction pairs.
			All dots with black borders depict the same image.
            \textbf{(a)} The output logits from two single models. We observe frequent occurrences of crossing the decision boundary, leading to either positive or negative flipping.
            Notably, test samples flip even if not close to the boundary, illustrated by {\color{magenta}magenta long arrows}.
            \textbf{(b)} The output logits from two 3-model ensembles. 
            We see fewer samples that are far from the boundary flip, illustrated by {\color{magenta} magenta shorter arrows}.
            \textbf{(c)} The output logits from two 3-model ensembles and their members in the same plot.
            We observe that individual members' logits, illustrated in lighter circles, center around the mode that is illustrated in darker circles.
			\textbf{(d)} Estimated probability mass function (PMF) of logit displacement between two single models or ensembles.
			The $x,y$-axes denote the two classes' logit displacement.
			The heatmap value denotes the estimated probability density.
			The ensemble's co-variance is significantly smaller than the single model.
			The figure is best viewed in color.
		}
		\label{fig:toy_example}
	\end{center}
    \vspace{-10pt}
\end{figure*}

\myparagraph{Approximating logit displacement norm with a few flipping-susceptible classes.}
In reality, the normal distribution in \cref{eq:clt_different,eq:clt_same} is by no means isotropic especially when the underlying models ($\vphi,\psi$) learn meaningful representation.
Conversely, as will be illustrated in \cref{fig:hist2d_clt} and \cref{sec:prob:toy_example}, this distribution is anisotropic enough such that the logit displacement norm can be approximated by the sum of logit difference at a few ``outstanding'' classes whose magnitudes are high:
\begin{align}
\label{eq:approximate_norm}
\| \vphi^\mathrm{(ens)}(x) - \vpsi^\mathrm{(ens)}(x) \|_p^p
\approx \sum_{k\in\sK(x)} \left( | \vphi^\mathrm{(ens)}_k(x) - \vpsi^\mathrm{(ens)}_k(x) |^p \right),
\end{align}
where $ \sK(x) \subset \{1,\cdots, C\} \text{ such that } \forall j\in\sK(x) \text{ and } j' \in \{1,\cdots, C\} \setminus \sK(x), \vphi^{(\mathrm{ens})}_j(x) \geq \vphi^{(\mathrm{ens})}_{j'}(x) $.

\begin{wrapfigure}{r}{0.64\linewidth}
	\centering
	\includegraphics[width=\linewidth]{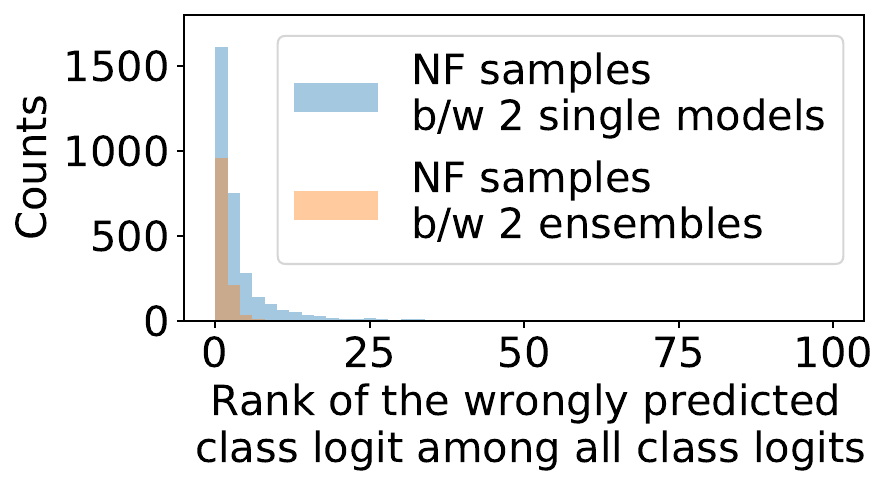}
	\caption{
    \textbf{Rank of negative flips' wrong prediction in the old model.}
    The wrongly predicted class of the negative flip samples in the new model also ranks high in the old model, motivating us to focus on those flipping-susceptible classes.
    The experiments are done on ImageNet-1K and $C=1000$.}
	\label{fig:logit_rank}
\end{wrapfigure}

Furthermore, not coincidentally, this subset of classes with high logit values is more prone to prediction flipping.
We take negative flipped samples between two ResNet-18 models and plot the rank of the wrongly predicted class's logit among all logits in the other model, namely $\mathrm{Rank}(\vpsi_{\arg\max(\vphi(x))}(x), \vpsi(x))$, in \cref{fig:logit_rank}.
We can see that these negative flips' wrong prediction will also rank high, although not necessarily the topmost, in the other model that classifies it correctly.
This phenomenon becomes more prominent if two models are ensemble. 
It motivates the formulation in \cref{eq:ldi}.

\begin{figure*}[!tb]
    \centering
    \subfloat[
    ]{
        \includegraphics[width=0.48\linewidth]{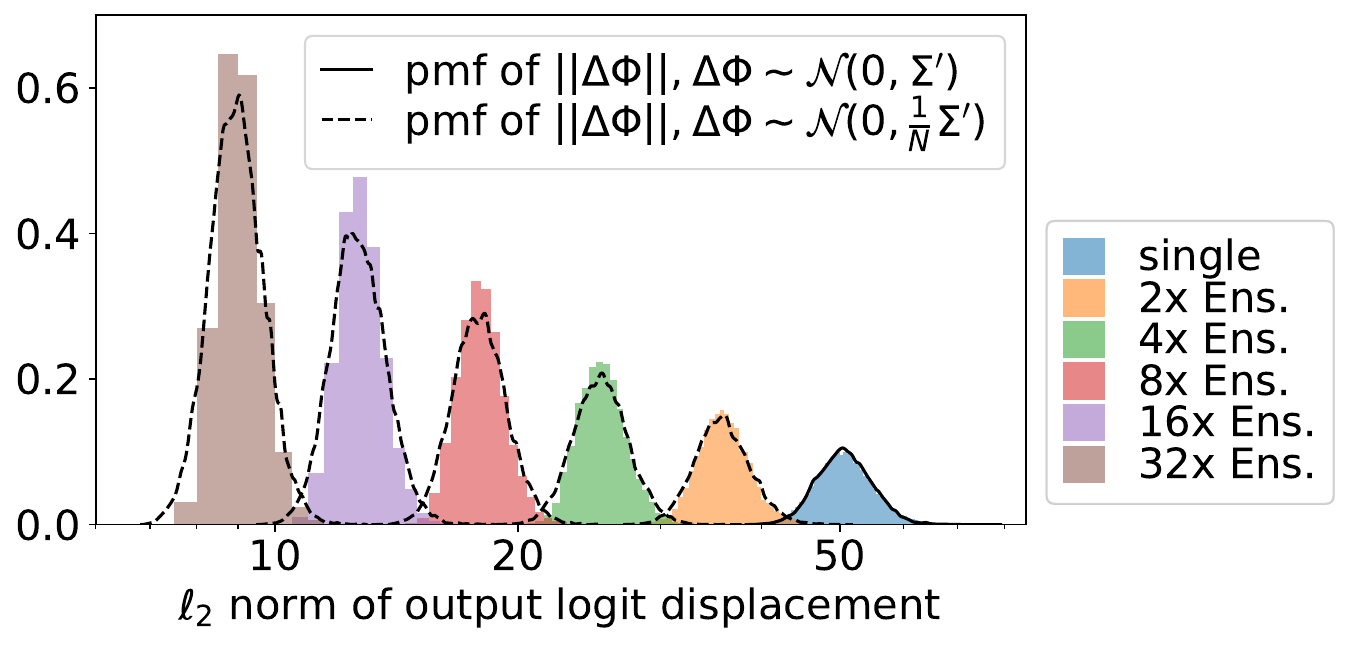}
        \label{fig:hist_high_dim:logit_diff}
    }
    \subfloat[
    ]{
        \includegraphics[width=0.48\linewidth]{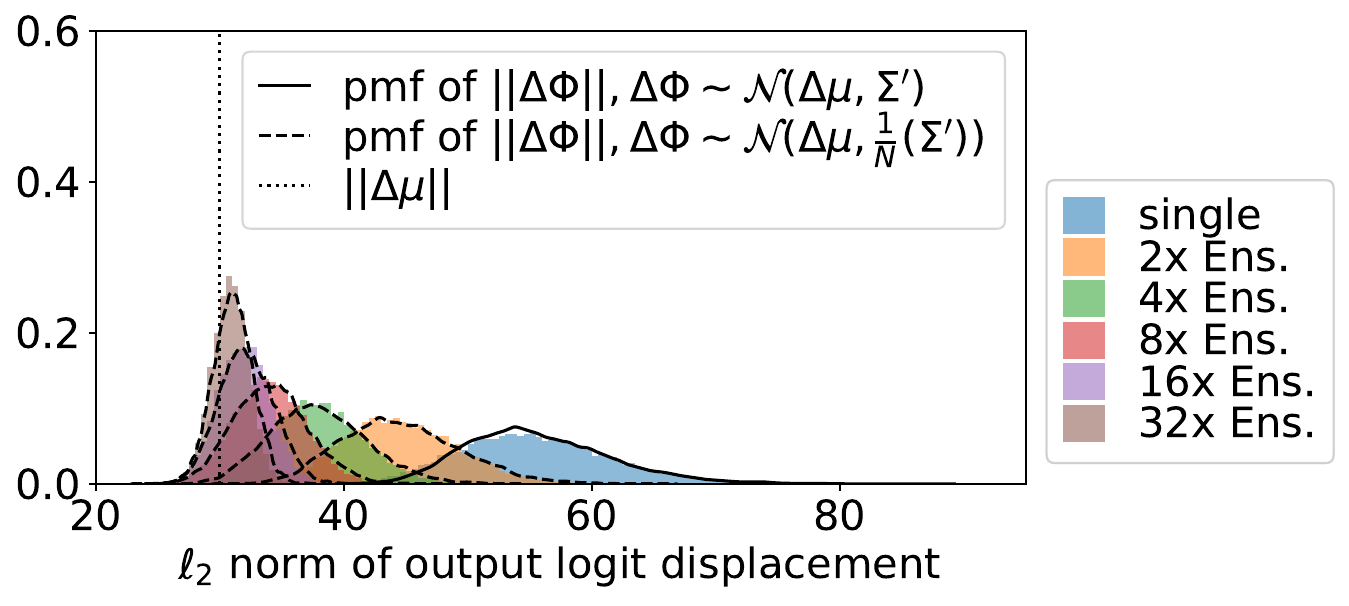}
        \label{fig:hist_high_dim:logit_diff_hetero}
    }
    \caption{
            \textbf{$\ell_2$ norm histogram of logit displacement between two random ensembles.}
            The bin size is $0.5$.
            We also plot the simulated PMFs:
            solid lines for $\ell_2$ norm of a simulated normal distribution $\mathcal{N}\left(\Delta\vmu, (\mSigma_1+\mSigma_2)\right)$ whose parameters are estimated from all available single models;
            dashed lines for those of extrapolated distribution $\mathcal{N}\left(\Delta\vmu, \frac{1}{m}\mSigma'\right)$.
            Consistency between the histograms and PMFs supports our hypotheses in \cref{sec:probe:landscape}.
            \textbf{(a)} denotes the case of a homogeneous ResNet-18 ensemble and another ResNet-18 homogeneous ensemble. In this case, we have $\Delta\vmu=\vzero, \mSigma'=2\mSigma_{1,2}$.
            \textbf{(b)} denotes the case of a homogeneous ResNet-18 ensemble and a ResNet-50 homogeneous ensemble. In this case, we have $\Delta\vmu\neq\vzero, \mSigma'=\mSigma_1 + \mSigma_2$.
    }%
    \label{fig:hist_high_dim}
\end{figure*}

\subsection{A Two-Dimensional Example}
\label{sec:prob:toy_example}

To illustrate the behavior of models in logit space, we create a toy example by selecting {\em two} classes, ~\ie``Labrador retriever'' (n02099712) and ``French bulldog'' (n02108915), from ImageNet~\cite{deng2009imagenet} and training several ResNet-18 models for {\em binary} classification. The models differ in their 
initialization, determined by distinct random seeds; we then collect output logits for each test datum and model in the ensemble. 
In \cref{fig:toy_model_single}, we plot the two-dimensional logit vectors of multiple data points when updating from one individual model to another. We can roughly categorize the negative flipped samples, highlighted with the purple arrows, into two types: (1) those close to the decision boundary in the old model; and (2) those far from the decision boundary in the old model but still flipped in the new one, due to significant displacement of the logit vector. 
\cref{fig:toy_model_ensemble} shows the logit vectors of the same set of data points but in the update case of two ensemble models each having 3 members ($3\times$).
Compared to~\cref{fig:toy_model_single}, we can observe a clear reduction in the magnitude of displacement during the update.
To validate that this observation is not incidental, we construct many cases of model updates and measure the distribution of the logit vector displacement on a certain data sample. As shown in~\cref{fig:hist2d_clt}, in updates between ensembles, the logit vectors are less likely to exhibit significant displacement. We repeat the same visualization on more data points in~\cref{appendix:visualization} to come.
This suggests that the ensemble may be reducing the negative flip rate through the reduction of displacements of the logit vectors.

\subsection{Validating the Hypothesis of Reducing Logit Displacement Magnitude}
\label{sec:probe:val}

\noindent We validate our hypothesis on the representation landscape through large-scale experiments.
Specifically, we train 256 ResNet-18 models on ImageNet with different seeds
and split them into two halves.
For an arbitrary image, we randomly draw $m$ models without replacement in each half and compute the averaged logits of this drawn ensemble.
We repeat the process and present in \cref{fig:hist_high_dim:logit_diff} the histogram of the logit displacement's $\ell_2$ norm between two random ensembles.
Note that Seguin~\etal~\cite{seguin2021understanding} argue that the logit distribution is highly affected by the number of training epochs, therefore we follow the standard training recipe, which is detailed in \cref{sec:exp:setup}, for the assumption to hold.

We examine our hypothesis in \cref{sec:probe:landscape} by comparing the histogram with the probability mass function (PMF) of the logit displacement norm.
We start by using all available single models to estimate mean and co-variance $(\vmu,\mSigma)$ for the logit vectors' distribution $\vphi(x)$ and examine whether the norm of logit displacement will follow a generalized (central) $\chi^2$ distribution.
Since the probability density function (PDF) of a generalized $\chi^2$ variable does not have a simple closed-form expression, we approximate it by Kernel Density Estimation (KDE)~\cite{parzen1962kde}.
From \cref{fig:hist_high_dim:logit_diff}, we see that the simulated PMFs in solid lines fit the histogram of single models well, implying that logits of these models could indeed follow a normal distribution.
We conducted the same experiments above on more images in~\cref{appendix:visualization} and the conclusion holds well, suggesting this property is not incidental.

We also present a theoretical argument for the sample-wise output logit approximating a normal distribution.
We assume the output logit $y = \vphi(x; w)$ to be locally linear to the parameters $w$ given an arbitrary sample $x$.
Since $w$ are initialized with a normal distribution,~\eg Xavier~\cite{glorot2010xavier} or Kaiming~\cite{he2015kaiming} initialization, $\vphi(x; w)$ will locally follow another normal distribution at initialization.
The lazy-training theory~\cite{chizat2019lazy} hypothesizes that a deep model behaves as its linearization around the initialization.
If it holds, we can derive that $\vphi(x; \hat{w})$, where $\hat{w}$ denotes the learned weights after training, again follows a normal distribution.

If we move to the ensemble case, the logit displacement follows another normal distribution with a scaled covariance matrix,
which leads to another generalized $\chi^2$ distribution with shifted mean and scaled covariance.
Its estimated PMF, shown in dashed lines in \cref{fig:hist_high_dim:logit_diff}, is consistent with the corresponding histogram.

The sample-specific logit displacement norm between two homogeneous ensembles with different architectures can be analyzed likewise.
Considering an arbitrary image $x$,  $\vphi(x)\sim\Normal(\vmu_1,\mSigma_1)$ for ResNet-18 and $\vpsi(x)\sim\Normal(\vmu_2,\mSigma_2)$ for ResNet-50, the norm of logit displacement should follow a generalized \emph{non-central} $\chi^2$ distribution.
From \cref{fig:hist_high_dim:logit_diff_hetero} we can see that the estimated distribution of the logit displacement norm fits the empirical distribution well.
It still condenses as the ensemble gets larger but much slower than \cref{fig:hist_high_dim:logit_diff}.
Also, the mean converges to non-zero (the dotted vertical line in \cref{fig:hist_high_dim:logit_diff_hetero}).

\section{Method: Ensemble Logit Difference Inhibition (\elodi)}
\label{sec:method}

\noindent The above analysis suggests the effectiveness of large homogeneous ensembles in reducing NFR, but an ensemble is less practical compared with a single model due to its multiplied inference cost to run every member model on a new input.
In this work, we propose to re-purpose the knowledge distillation technique~\cite{hinton2015kd}, which was previously used for improving model accuracy, for transferring the NFR reduction capability from ensembles to a single model.

\subsection{Updates with Ensemble-Distilled Models}
\label{sec:method:elodi}

\noindent Given an ensemble $\vphi^\mathrm{(ens)}$ composed of a set of models $\sM = \{\vphi^{(i)}\}_{i=1}^m$, we learn a {\em single} model $\hat{\vphi}$ such that for each sample $x\in\sX$, the random variable of sample logits has ensemble-like reduced variance. 
We then use single models learned in this way in every model update of a system.

As illustrated in \cref{fig:teaser}(a),
when an old model $\hat\vphi^\mathrm{(old)}$ needs to be updated to a new model, we first train a homogeneous ensemble on the same dataset and having the same architecture as the desired new model.
Next, we distill this reference ensemble to the actual new model $\hat\vphi^\mathrm{(new)}$.
If the distillation process can convey the property of reducing NFR to the learned single model, updating from $\hat\vphi^\mathrm{(old)}$, which was preferably produced in the same manner, to $\hat\vphi^\mathrm{(new)}$ would result in significantly reduced NFR than that of a model pair without this treatment.

We empirically verify this model update method, called \elodi, in \cref{sec:exp:main} and find it effective in reducing NFR while retaining the accuracy gain introduced by the new architecture used in the new model.
It avoids the prohibitive inference cost of deep ensembles due to only using $\hat\vphi^\mathrm{(new)}$ in the model update, with reference ensemble discarded after the distillation.
Another benefit of \elodi is that the new model does not need to target any specific existing model in an update, enabling a chain of models to yield relatively low NFR between any pair of them. This is rather helpful when multiple updates are consecutively executed, which is common for a long-running machine learning system. 
The overall pipeline of a sequential model update via \elodi is summarized in \cref{alg:elodi}.

\begin{algorithm}[tb]
	\caption{Sequential model update with \elodi}
	\label{alg:elodi}
	\begin{algorithmic}
		\STATE {\bfseries Input:} Dataset $\sX$, number of versions to update $T$
		\STATE {\bfseries Output:} A sequence of models $\hat\vphi^{(\mathrm{Ver\mhyphen}0)}, \cdots, \hat\vphi^{(\mathrm{Ver\mhyphen}T)}$.
		\STATE $t \gets 0$  %
		\WHILE{$t \leq T $}
		\FOR{$i=1$ {\bfseries to} $m$}  %
		\STATE $\vphi^{(i,\mathrm{Ver\mhyphen}t)}\gets{\arg\min}_{\vphi^{(\mathrm{Ver\mhyphen}t)}}\sum_{x\in\sX}\loss_\mathrm{CE}\left(\vphi^{(\mathrm{Ver\mhyphen}t)}\right)$
		\ENDFOR
		\FORALL{$x \in \sX$}  %
		\STATE $\vphi^{(\mathrm{ens},\mathrm{Ver\mhyphen}t)}(x)\gets\frac{1}{m} \sum_{i=1}^{m}\vphi^{(i,\mathrm{Ver\mhyphen}t)}(x)$
		\ENDFOR
		\STATE {\footnotesize $\hat\vphi^\mathrm{(\mathrm{Ver\mhyphen}t)}\gets\arg\min_{\vphi} \sum_{x\in\sX} \loss_\mathrm{distill} \left(\vphi(x), \vphi^{(\mathrm{ens},\mathrm{Ver\mhyphen}t)}(x) \right)$ }
		\STATE $t \gets t + 1$  %
		\ENDWHILE
	\end{algorithmic}
\end{algorithm}

\subsection{Loss Choices of \elodi}
\label{sec:method:distill_loss}

\noindent Generally we obtain the model to be deployed using the distillation technique~\cite{hinton2015kd}, 
\begin{align}
	\label{eq:optimize_distill}
	\hat\vphi = \underset{\vphi}{\arg\min} \sum_{x\in\sX} \loss_\mathrm{distill}
	\left(\vphi(x); \vphi^\mathrm{(ens)}(x) \right).
\end{align}

Various types of distilling functions have been proposed to improve single-model accuracy: (1) Vanilla KD loss~\cite{hinton2015kd} minimizes the KL-divergence between two models' output logits; (2) FitNet~\cite{romero2015fitnet} and AttentionTransfer~\cite{zagoruyko2017paying} mimics the intermediate hidden layer's activation or attention; (3) FSP~\cite{yim2017gift} mimics cross-layer Gram matrices.

We empirically find that applying exact logit matching, which minimizes the $\ell_p$-norm of logit difference from the single model to a reference ensemble model, achieves the goal of distilling reduced variance from an ensemble,
\begin{align}
	\label{eq:logit_matching}
	\loss_\mathrm{distill}(x) 
	= \sum_{k=1}^C
	\left( \| \vphi_k(x) - \vphi^{\mathrm{(ens)}}_k(x) \|_p\right)^p.
\end{align}

Furthermore, due to the fact stated in the last paragraph of \cref{sec:probe:landscape} that the logit displacement magnitude is dominated by a few elements with a large difference in the high-dimensional case, we propose a top-$K$ variant of \cref{eq:logit_matching}, called Logit Difference Inhibition (LDI) loss, which only inhibits logit difference on those classes with the top-$K$ highest logit elements.
\begin{align}
	\label{eq:ldi}
	\loss_\mathrm{LDI}(x) =
	\sum_{k\in\sK(x)}
	\left( \| \vphi_k(x) - \vphi^{\mathrm{(ens)}}(x) \|_p\right)^p,
\end{align}
where $ \sK(x) $ follows the definition in \cref{eq:approximate_norm} or $\sK(x) = \texttt{np.argsort(}-\vphi_k(x)\texttt{)[0:K]}$ in pythonic pseudocode.

As will be shown in experiments, the Top-$K$ formulation leads to no loss in NFR reduction compared to the full form.
It could instead help in reducing computation cost when the number of classes are extremely large~\cite{an2020partialfc}.
Most importantly, it implies that \elodi transfers the capability of NFR reduction in ensembles by reducing the variance of logit estimation indeed instead of exact logit matching through $\ell_p$ norm.
$ p $ is 2 in our experiments. $ K $ is 10 by default according to \cref{fig:logit_rank}.

\myparagraph{The overall objective} of the distillation in \elodi is a weighted sum of standard Cross-Entropy and the LDI loss,~\ie $ \loss = (1-\alpha)\loss_\mathrm{CE} + \alpha\loss_\mathrm{LDI} $,
where the weight $ \alpha $ is set such that the magnitude of $\loss_\mathrm{CE}$ and $\loss_\mathrm{LDI}$ is similar.

\subsection{Integrating \elodi with existing models}
\label{sec:method:integrate}

\noindent Dealing with old models without \elodi is necessary when updating an existing system. 
We consider the simple case of one old model not trained with \elodi.
In this case, we augment \elodi with an additional LDI loss \wrt to the old model,~\ie
\begin{align}
	\loss_\mathrm{total} &= \lambda\loss_\mathrm{LDI}(\model^\diamondsuit_\mathrm{new}; \model^\mathrm{(ens)}_\mathrm{new}) + (1-\lambda)\loss_\mathrm{LDI}(\model^\diamondsuit_\mathrm{new}; \model_\mathrm{old}),
\end{align}
where $\model^\diamondsuit$ denotes the model to be learned from the ensemble.

\section{Experiments}
\label{sec:exp}

We conduct extensive experiments to showcase the effectiveness of the proposed \elodi from multiple aspects.
In~\Cref{sec:exp:setup}, we describe the dataset statistics, evaluation metrics, and implementation details of training.
In~\Cref{sec:exp:main}, we show the main results of \elodi under the standard setting of updating a ResNet-18 model to a ResNet-50 model in comparison with other existing positive-congruent training (PC-Training) methods.
In~\Cref{sec:exp:more}, we demonstrate that \elodi is effective on more practical settings of model updates, including (1) the data-growth setting where more data is available for training the new model, (2) more than two rounds of model updates, (3) integrating \elodi with existing models, and (4) updating the old model to new models with various architectures.
In~\Cref{sec:exp:elodi_ablation}, we study the design choices in \elodi.
Particularly, we show the advantages of using homogeneous ensembles as guidance instead of heterogeneous ones in \elodi to  support the analysis in~\Cref{sec:probe:landscape}.
In~\Cref{sec:exp:elodi_more_ablations}, we provide some exploratory studies on the parameters in \elodi.

\subsection{Experimental Setup}
\label{sec:exp:setup}

\myparagraph{Datasets.}
We validate the proposed approaches on two standard image classification datasets: ImageNet~\cite{deng2009imagenet} and iNaturalist~\cite{van2018inaturalist}.
For ImageNet, we use ILSVRC12~\cite{russakovsky2015imagenet} which contains 1,000 categories.
It has around 1.2 million training images and 50,000 validation images.
For iNaturalist, we use the version released in 2017.
It covers 5,089 categories of fine-grained species, with 579,184 training images and 95,986 validation images.
Besides, to showcase the efficacy of \elodi on more data modalities, we also conduct experiments on the AG News Corpus~\cite{zhang2015character}, a text classification dataset which contains 4 classes (``World'', ``Sports'', ``Business'', ``Sci/Tech'') of news articles.
It has 30,000 training and 1,900 test samples for each class.

\myparagraph{Metrics.}
In a model update experiment, we measure the top-1 error rate (\textbf{ER}) of both the old and new models.
{\small
\begin{align}
	\mathrm{ER}_\mathrm{old} = \frac{1}{N}\sum_{i=1}^N \mathbb{I}\left(\hat{y}_i^\mathrm{(old)} \neq \ell_i\right),
	\mathrm{ER}_\mathrm{new} = \frac{1}{N}\sum_{i=1}^N \mathbb{I}\left(\hat{y}_i^\mathrm{(new)} \neq \ell_i\right).
\end{align}
}
In addition, we report the negative flip rate with respect to the old model (\textbf{NFR})~\cite{yan2021pct}, which is defined as
\begin{align}
\mathrm{NFR} &= \frac{1}{N}\sum_{i=1}^N \mathbb{I}\left(\hat{y}_i^\mathrm{(new)} \neq \ell_i, \hat{y}_i^\mathrm{(old)} = \ell_i\right),
\end{align}
where $\mathbb{I}(\cdot)$ is the indicator function, $\ell_i$ is the label,  and $\hat{y}_i^\mathrm{(new)}$ ($\hat{y}_i^\mathrm{(old)}$) is the new (old) model's prediction.

Since the absolute value of NFR is upper bounded by the error rate, comparing methods with different accuracies is hard.
To this end, we also report \textbf{Relative NFR} (\textbf{Rel-NFR})~\cite{yan2021pct}, which is defined as
\begin{align}
	\mathrm{Rel\mhyphen NFR} &= \frac{\mathrm{NFR}}{(1-\mathrm{ER}_\mathrm{old})\times\mathrm{ER}_\mathrm{new}},
\end{align}
where the denominator is the expected error rate on the subset of samples predicted correctly by the old model.
Relative NFR factors out overall model accuracies.

\myparagraph{Implementation Details.}
We follow the standard training recipe of ImageNet\footnote{\url{https://github.com/pytorch/examples/tree/main/imagenet}.}.
Specifically, all classification models are trained with SGD with a momentum of 0.9 and a base learning rate of 0.1, which is reduced by a tenth every 30 epochs until 90 epochs.
The batch size is 256 with 8 GPUs.

For \elodi with larger models and ensemble size, GPU memory becomes a bottleneck.
To handle the memory issue, we use gradient checkpointing~\cite{chen2016training} and reduce batch size while linearly scaling the base learning rate~\cite{goyal2017accurate}.

Unless otherwise specified, \elodi experiments are done with ensemble size $m=8$.

\subsection{Main Results of \elodi}
\label{sec:exp:main}

\begin{table}[t]
	\centering
	\caption{
		\textbf{Comparing \elodi with other PC-Training Methods.}
		$^*$ in \hl{ensemble paragon} means that the number is from a collection of models.
		The middle rows are targeted model update baselines.
		The bottom rows are \elodi and its Top-$K$ variants.
	}
	\label{tab:main_result}
	\setlength{\tabcolsep}{0.5em}
    \resizebox{\linewidth}{!}{
	\begin{tabular}{l|c|c|c|c}
		\toprule
		\multirow{2}{*}{Method} & \multicolumn{2}{c|}{ER\lowbetter (\%)}   & NFR\lowbetter (\%) & Rel-NFR\lowbetter(\%) \\
		& RN-18 & RN-50 & & \\
		\cmidrule{1-5}
		No treatment (single)    & 30.24 & 24.66 & 4.30 & 25.00 \\
		\rowcolor{Gray}
		Ensemble Paragon ($8\times$) & ~~26.34$^*$ & ~~22.44$^*$ & 1.95 & 11.80 \\
        Dropout~\cite{srivastava2014dropout} & 30.97 & 24.50 & 4.08 & 21.92 \\
		\cmidrule{1-5}
		\multicolumn{4}{l}{\textsc{(Targeted model update)}} & \\
		BCT~\cite{shen2020bct} & 30.24 & 25.00 & 4.34  & 24.66 \\
        RACT~\cite{zhang2022ract} & \ditto & 26.56 & 3.35 & 18.08 \\
		KD~\cite{hinton2015kd} & \ditto & 28.38 & 3.20 & 16.16 \\
		FD-KL~\cite{yan2021pct} & \ditto & 26.32   & 2.90  & 15.79 \\
        FD-LM~\cite{yan2021pct} & \ditto & 26.53   & 2.92  & 15.78 \\
		BU-CR~\cite{trauble2021backward} & \ditto & 26.51 & 4.56 & 24.66 \\
		\cmidrule{1-5}
		\multicolumn{4}{l}{\textsc{(Ours)}} & \\
		\elodi ($K=1000$) & 31.34 & 23.15 & 2.18 &  13.72 \\ 
		\elodi ($K=10$) & 30.95 & \textbf{23.10}  & \textbf{2.11} & \textbf{13.23} \\
		\bottomrule
	\end{tabular}
    }
\end{table}

\myparagraph{Comparison with other PC-Training Methods.}
We first compare \elodi with other PC-Training methods under the setting of updating ResNet-18 to ResNet-50 in~\cref{tab:main_result}.
``No treatment'' means that both models are trained with standard cross entropy loss.
Previous methods include (a) Backward-Compatible Training (BCT)~\cite{shen2020bct}, (b) Regression-Alleviating Compatible Training (RACT)~\cite{zhang2022ract}, (c) Knowledge Distillation (KD)~\cite{hinton2015kd}, (d) Focal Distillation with either KL divergence or Logit Matching as the objective (FD-KL/LM)~\cite{yan2021pct}, and (e) Bayesian Update (BU)~\cite{trauble2021backward}\footnote{In our re-implementation, we use CostRatio (CR)-based Bayesian update rule to get the target and train the new model online (see discussion in~\cref{sec:exp:elodi_more_ablations} and~\cref{tab:offline_distillation}) instead of offline model update in the original paper.}.
BCT and RACT belong to compatible-training-based methods while KD and FD-KL/LM are distillation-based methods.
We can see that \elodi outperforms previous methods in terms of both absolute and relative NFR.
Note that these baselines are all {\em targeted model update}, meaning that the old ResNet-18 is used as the target when training ResNet-50.
In contrast, \elodi does not target any legacy model.

Since Dropout~\cite{srivastava2014dropout} implicitly makes a network behave similar to an ensemble~\cite{gal2016dropout}, we also report its NFR performance using a dropout probability of $p=0.2$.
The NFR is still relatively high (4.08 \versus 4.30\%) partially due to the increased ER of both ResNet-18 and -50.
Though the relative NFR is smaller than the non-treatment baseline (21.92 \versus 25.00\%), it is still significantly greater than \elodi (13.23\%).
One reason is that the ensemble property of dropout is ``weak''. The other might be a considerable correlation among the hypothetical ``ensemble members'' in the dropout setting due to the shared internal representations hindering the reduction of NFR as ensembles.

\begin{table*}[!tb]
	\begin{center}        
		\caption{
				\textbf{\elodi on iNaturalist.}
                $^*$ in \hl{ensemble paragon} means that the number is from a collection of models.
                \half(\full)~means that the model is trained on half (full) data.
				\elodi is effective when fine-tuning on iNaturalist under multiple settings.
		}
		\label{tab:iNaturalist}
		\setlength{\tabcolsep}{0.5em}
		\begin{tabular}{c|c|c|c||c|c|c||c|c|c}
			\toprule
			\multirow{3}{*}{Method} & \multicolumn{3}{c||}{Increasing \#classes} & \multicolumn{3}{c||}{Increasing \#samples/class} & \multicolumn{3}{c}{Full data}\\
			\cmidrule{2-10}
			& \multicolumn{2}{c|}{ER\lowbetter (\%)} & \multirow{2}{*}{NFR\lowbetter} & \multicolumn{2}{c|}{ER\lowbetter (\%)} & \multirow{2}{*}{NFR\lowbetter} &
			\multicolumn{2}{c|}{ER\lowbetter (\%)} & \multirow{2}{*}{NFR\lowbetter} \\
			\cmidrule(lr){2-3}
			\cmidrule(lr){5-6}
			\cmidrule(lr){8-9}
			& RN-18\half & RN-50\full &  & RN-18\half & RN-50\full & &  RN-18\full & RN-50\full & \\
			\midrule
			No treatment   &  47.58 & 35.95  & 5.38  & 78.88 & 35.95 &  3.82 & 40.69 & 35.95 & 4.76 \\
			Focal Distillation~\cite{yan2021pct} & \ditto & 45.87 & 3.44 & \ditto & 66.91 & 2.00 & \ditto & 40.03 & 3.95 \\
			\rowcolor{Gray}
			Ensemble Paragon ($8\times$) &  ~~36.03$^*$ &  ~~29.47$^*$ & 1.68 & ~~47.04$^*$ & ~~29.47$^*$  &    1.23  & ~~36.25$^*$ & ~~29.47$^*$ & 2.10 \\ %
			\elodi        &    43.56    &    34.29    &  1.91   &    52.96   &   34.29    &   1.47  & 40.37 & 34.29 & 2.46 \\
			\bottomrule
		\end{tabular}
	\end{center}
    \vspace{-10pt}
\end{table*}

\begin{table}[!tb]
	\centering
	\caption{
			\textbf{\elodi in data-growth settings on ImageNet.}
            $^*$ in \hl{ensemble paragon} means that the number is from a collection of models.
			\half(\full)~means that the model is trained on half (full) data.}
	\label{tab:incremental}
	\setlength{\tabcolsep}{0.5em}
    \resizebox{\linewidth}{!}{
	\begin{tabular}{c|c|c|c||c|c|c}
		\toprule
		\multirow{3}{*}{Method} & \multicolumn{3}{c||}{Increasing \#classes} & \multicolumn{3}{c}{Increasing \#samples/class} \\
		\cmidrule{2-7}
		& \multicolumn{2}{c|}{Error Rate\lowbetter (\%)} & \multirow{2}{*}{NFR\lowbetter} & \multicolumn{2}{c|}{Error Rate\lowbetter (\%)} & \multirow{2}{*}{NFR\lowbetter} \\
		\cmidrule{2-3}
		\cmidrule{5-6}
		& RN-18\half & RN-50\full &  & RN-18\half & RN-50\full  &   \\
		\midrule
		No treatment     &   22.02     & 24.66  &   14.07  & 34.26  & 24.66 &  3.52    \\
		Focal Distillation~\cite{yan2021pct} &     \ditto   &  39.96 & 5.45 &   \ditto    & 33.06 & 2.65 \\ 
		\rowcolor{Gray}
		Ensemble Paragon ($8\times$)  & ~~18.70$^*$ & ~~22.44$^*$ & 4.12 &   ~~29.16$^*$    &  ~~22.44$^*$      &   2.11   \\
		\elodi        &  21.80 & 23.15 & 4.19 &    34.08   &   23.15    &  2.25   \\
		\elodi ($K=10$) &  21.24 & 23.10 & 4.10 &  33.93     &   23.10    &   2.21 \\
		\bottomrule
	\end{tabular}
    }
\end{table}

Finally, we study a few variants of \elodi.
Using the Top-$K$ highest-logit class subset in \elodi with $ K=10 $ even slightly outperforms distilling all classes of logits by 0.07\%.
From an ER-NFR scatter plot in \cref{fig:teaser}(b), \elodi achieves a similar level of ER-NFR results as the ensemble paragon~\cite{yan2021pct} at the inference cost of a single model.

\subsection{More Update Settings of \elodi}
\label{sec:exp:more}

In this subsection, we move beyond the standard comparison with other PC-Training methods and look at a few practical scenarios where model updates occur.
First, we study the case of model updates when the training data grows in either the number of classes or the number of per-classes samples.
Second, we show the applicability of \elodi on other datasets.
Third, we demonstrate that \elodi enables multiple rounds of model updates.
Fourth, we showcase that \elodi is seamlessly integrated with existing models that have been deployed without positive-congruent training. 
Finally, we reveal that the NFR reduction ability of \elodi is agnostic to architectures.

\myparagraph{Data-growth settings.}
Model updates may also come with growing training data,~\eg(1) increasing number of classes, and (2) increasing number of per-class samples.
We follow the same data/class split in~\cite{yan2021pct}, which uses 50\% classes/samples for the old model and full data for the new.
From~\cref{tab:incremental}, we find that conclusions from the full-data setting hold for these settings.
In particular, \elodi achieves a significantly lower error rate compared to Focal Distillation (FD) because \elodi is not targeted at the old model whose performance is typically much worse due to less training data.

\myparagraph{Fine-tuning on other datasets.} 
We validate the effectiveness of \elodi when transferring to a different dataset.
We examine iNatualist~\cite{van2018inaturalist} and follow the protocol in~\cite{yan2021pct}.
Results of both full-data and data-growth settings are given in~\cref{tab:iNaturalist}, where \elodi consistently outperforms FD.

\begin{table}[t]
	\centering
	\caption{
		\textbf{\elodi on textual data.}
        On the AG News Corpus, \elodi reduces the relative NFR between two text classification models by $68.8\%$ compared to the no-treatment baseline.
        The performance is close to the \hl{ensemble paragon}, where all numbers are obtained from a collection of models.
        Here, both the old and new models follow the BERT-base architecture~\cite{devlin2018bert}.
	}
	\label{tab:text_classification}
	\setlength{\tabcolsep}{0.5em}
    \resizebox{\linewidth}{!}{
	\begin{tabular}{l|c|c|c|c}
		\toprule
		\multirow{2}{*}{Method} & \multicolumn{2}{c|}{ER\lowbetter (\%)}   & NFR\lowbetter (\%) & Rel-NFR\lowbetter(\%) \\
		& old & new & & \\
		\cmidrule{1-5}
		No treatment (single)    & 5.95 & 5.95 & 0.934 & 16.69 \\
		\rowcolor{Gray}
		Ensemble Paragon ($8\times$) & ~~5.82$^*$ & ~~5.62$^*$ & ~~0.263$^*$ & ~~4.97$^*$ \\
		\cmidrule{1-5}
		\elodi & 5.90 & 5.89 & 0.289 & 5.21 \\ 
		\bottomrule
	\end{tabular}
    }
    \vspace{-10pt}
\end{table}

\myparagraph{\elodi on more data modalities.}
To show the efficacy of \elodi on other modalities, we conduct a set of text classification experiments on the AG News corpus.
Particularly, we consider the model update between two BERT-base models~\cite{devlin2018bert}.
Both old and new models take the pre-trained weights and are then fine-tuned on the AG News corpus, following~\cite{sun2019fine}.
The results are shown in~\cref{tab:text_classification}.
We can see that \elodi reduces both absolute and relative NFR by almost $70\%$ compared to the no-treatment baseline.

\begin{table}[!tb]
	\centering
	\caption{
			\textbf{\elodi achieves lower Pairwise NFR on a chain of models.}
			$\model_1\rightarrow\model_2$ means that we measure $\model_2$'s NFR \wrt $\model_1$.
			$\model_{1}\Rightarrow\model_{2}$ means that $\model_2$ is trained with $\model_{1}$ being teacher using distillation loss,~\eg FD.
			$\model^\diamondsuit$ means that $\model$ is trained from \elodi.
			(1) \textbf{chain}: each model targets at its closest predecessor;
			(2) \textbf{radial}: each model targets at its farthest ancestor;
			(3) fully connected (\textbf{fc}): each model targets at all its ancestors.
	}
	\label{tab:architecture_chain}
	\setlength{\tabcolsep}{0.5em}
	\begin{tabular}{c|c}
		\toprule
		Method  & Pairwise NFR  \\
		\midrule
		None &
		\begin{tikzcd}
			\text{RN-18} \arrow[r,"4.28\%"'] \arrow[rr, "3.92\%"', bend left=24] & \text{RN-50} \arrow[r,"4.41\%"'] & \text{RN-101}
		\end{tikzcd} \\
        \cmidrule{1-2}
        FD~\cite{yan2021pct} (chain) &
		\begin{tikzcd}
			\text{RN-18} \arrow[r,"2.90\%"', Rightarrow] \arrow[rr, "3.46\%"', bend left=24] & \text{RN-50} \arrow[r,"2.13\%"',Rightarrow] & \text{RN-101}
		\end{tikzcd}  \\
		FD~\cite{yan2021pct} (radial) &
		\begin{tikzcd}
			\text{RN-18} \arrow[r,"2.90\%"', Rightarrow] \arrow[rr, "2.63\%"', bend left=24, Rightarrow] & \text{RN-50} \arrow[r,"2.33\%"'] & \text{RN-101}
		\end{tikzcd} \\
		FD~\cite{yan2021pct} (fc) &
		\begin{tikzcd}
			\text{RN-18} \arrow[r,"2.90\%"', Rightarrow] \arrow[rr, "2.96\%"', bend left=24, Rightarrow] & \text{RN-50} \arrow[r,"\textbf{1.97\%}"', Rightarrow] & \text{RN-101}
		\end{tikzcd} \\
        \cmidrule{1-2}
		\elodi & 
		\begin{tikzcd}
			\text{RN-18\distill} \arrow[r,"\textbf{2.04\%}"'] \arrow[rr, "\textbf{2.19\%}"', bend left=24] & \text{RN-50\distill} \arrow[r,"\textbf{2.25\%}"'] & \text{RN-101\distill}
		\end{tikzcd}  \\
		\bottomrule
	\end{tabular}
\end{table}

\begin{table*}[!tb]
	\centering
	\caption{
		\textbf{\elodi with different architectures on ImageNet}.
		\elodi effectively reduces NFR on a wide range of architectures.
		$^*$ in \hl{ensemble paragon} means that the number is from a collection of models.
		$^\dagger$ is obtained by our reproduction with different augmentation and training schedules from the official one.
		Note that all new models' NFR is measured \wrt ResNet-18 listed in the leftmost column.
	}
	\label{tab:architecture_change}
	\setlength{\tabcolsep}{0.5em}
	\begin{tabular}{c|c||c|c||c|c||c|c}
		\toprule
		& ER$_\downarrow$ (\%) &  ER$_\downarrow$ (\%) & NFR$_\downarrow$ (\%) &  ER$_\downarrow$ (\%) & NFR$_\downarrow$ (\%) & ER$_\downarrow$ (\%) & NFR$_\downarrow$ (\%) \\
		& RN-18 (old) & RN-101 & $\rightarrow$RN-101 & DenseNet-161 & $\rightarrow$DenseNet-161 & Swin-Tiny & $\rightarrow$Swin-Tiny \\
		\midrule
		None (single)           & 30.24  &  24.66  &  3.64 & 21.82 & 3.73  & ~~20.40$^\dagger$ & 3.77 \\
		\rowcolor{Gray}
		Ensemble Paragon ($8\times$)  & ~~26.34$^*$  & ~~20.05$^*$  &  1.72 & ~~18.90$^*$ & 2.06 & ~~18.37$^*$ & 2.60 \\
		\elodi       & 31.34 & 21.09 & 2.19  & 21.74 & 2.57 & 19.84 & 2.95 \\
		\bottomrule
	\end{tabular}
    \vspace{-10pt}
\end{table*}

\myparagraph{\elodi on a chain of model updates.}
As discussed in \cref{sec:method:elodi}, \elodi does not involve the guiding ensembles at the inference stage.
Nor does it target any legacy model.
Only the single model trained with \elodi is deployed to replace the old model, which is also trained with \elodi.
When the number of updates increases, this naturally forms a chain of models having low NFR between them, inducing a {\em transitive} reduced NFR.

We illustrate this transitivity of NFR reduction in a chain of updates of three models,~\ie ResNet-18$\rightarrow$ResNet-50$\rightarrow$ResNet-101.
In~\cref{tab:architecture_chain}, with \elodi, NFRs between the three models reduce to $2.04\%\unsim2.25\%$ from $3.92\%\unsim4.41\%$ (a relative reduction of $44.1\%\unsim52.3\%$), outperforming all previous methods, including three variants of FD~\cite{yan2021pct}:
(1) chain, where each model targets at its closest predecessor; (2) radial, where each model
targets at its farthest ancestor; (3) fully connected (fc): each model targets at all its ancestors.
Note this is achieved without crafting the complex reference schemes that are necessary for the FD baselines since all baselines require old models' references.

\begin{table}[!tb]
	\centering
	\caption{
		\textbf{Integrating \elodi with existing models.}
		\elodi can reduce NFR \wrt old models that have been deployed without \elodi by being jointly optimized with targeted distillation (denoted by $\model_{1}\Rightarrow\model_{2}$).
	}
	\label{tab:bridge_LDI_EnsLDI}
	\setlength{\tabcolsep}{0.9em}
	\begin{tabular}{c|c}
		\toprule
		LDI usage  & Pairwise NFR \\
		\cmidrule{1-2}
		None & 
		\begin{tikzcd}
			\text{RN-18} \arrow[r,"2.98\%"'] \arrow[rr, "2.56\%"', bend left=18] & \text{RN-50\distill} \arrow[r,"2.25\%"'] & \text{RN-101\distill}
		\end{tikzcd} \\
		\cmidrule{1-2}
		Once & 
		\begin{tikzcd}
			\text{RN-18} \arrow[r,"\textbf{2.85\%}"', Rightarrow] \arrow[rr, "2.56\%"', bend left=18] & \text{RN-50\distill} \arrow[r,"2.14\%"'] & \text{RN-101\distill}
		\end{tikzcd} \\
		\cmidrule{1-2}
		Both & 
		\begin{tikzcd}
			\text{RN-18} \arrow[r,"\textbf{2.85\%}"', Rightarrow] \arrow[rr, "\textbf{2.48\%}"', bend left=18, Rightarrow] & \text{RN-50\distill} \arrow[r,"\textbf{2.12\%}"'] & \text{RN-101\distill}
		\end{tikzcd} \\
		\bottomrule
	\end{tabular}
\end{table}

\myparagraph{Integrating \elodi with existing models.}
As discussed in \cref{sec:method:integrate}, \elodi is compatible with an existing system where old models have been deployed without \elodi.
To show some empirical results, we consider three models, ResNet-18$\rightarrow$ResNet-50$\rightarrow$ResNet-101, where the ResNet-18 model is trained without \elodi.
Therefore both ResNet-50\distill~and -101\distill~with \elodi treatment will have a higher NFR compared with ResNet-18.
To handle this, we introduce an additional LDI loss targeted at ResNet-18 when training ResNet-50 and (or) ResNet-101 using \elodi.
The results are shown in \cref{tab:bridge_LDI_EnsLDI}.
We can see that \elodi w/. LDI outperforms \elodi w/o. LDI on all pairwise NFRs, indicating that a simple fix by augmenting \elodi with LDI loss towards the existing model is effective in dealing with this legacy case.

\myparagraph{Updates to dissimilar architectures.}
\label{sec:exp:more_arch}
All of the above experiments are based on the ResNet-family models.
In real applications, however, model updates often occur because of architectural improvement.
In \cref{tab:architecture_change}, we study if \elodi is applicable to updates across dissimilar architectures besides similar ones (ResNet-101).
For dissimilar architectures, we take two more representative architectures, namely DenseNet and Tiny-Swin Transformer, as examples.
We see that \elodi effectively reduces NFR in all cases, with retained or sometimes decreased ER.

\subsection{Design Choices of \elodi}
\label{sec:exp:elodi_ablation}

\begin{table}[!tb]
	\begin{center}
		\caption{
			\textbf{\elodi with different guiding ensembles.}
			We consider ResNet-18 $\rightarrow$ResNet-50 via \elodi with an $8\times$-model ensemble.
			\textbf{All-diff-weak} (\textbf{-strong}) ensemble is composed of 8 different weak (strong) models with Top-1 Acc. $\approx 69\%$ ($75\%$) on ImageNet.
			The model list can be found in~\cref{tab:model_list}.
			\textbf{Mixed-weak} (\textbf{-strong}) ensemble is a mixture of 4$\times$ ResNet-18 and 4$\times$ VGG-13 (4$\times$ ResNet-50 and 4$\times$ DenseNet-121).
			$^\ddagger$ means that the models in each ensemble are calibrated~\cite{guo2017calibration}.
		}
		\label{tab:ensemble_type}
		\setlength{\tabcolsep}{0.5em}
		\begin{tabular}{l|l|c|c|c}
			\toprule
			\multirow{2}{*}{Old Reference} &  \multirow{2}{*}{New Reference} & \multicolumn{2}{c|}{Error Rate$_\downarrow$ (\%)}   & NFR$_\downarrow$ (\%) \\ %
			\cmidrule(lr){3-4}
			&   & RN-18\distill & RN-50\distill  \\ %
			\midrule
			N/A & N/A               & 30.24 & 24.66 & 4.30 \\ %
			\midrule
			All-diff-weak & All-diff-weak  & 32.38 & 26.11 &  1.94\\ %
			Mixed-weak & Mixed-weak & 32.75  & 26.88  & 1.99 \\
			RN-18 ($\times8$) & RN-18 ($\times8$) & 31.32 & 26.82 & 2.06\\ %
			\midrule
			All-diff-weak & All-diff-strong  & 32.73 & 23.14 &  2.25\\ %
			All-diff-weak$^\ddagger$ & All-diff-strong$^\ddagger$ &  33.38 & 23.33 & 2.35 \\
			Mixed-weak & Mixed-strong & 32.75 & 23.68 & 2.24 \\
			RN-18 ($\times8$) & RN-50 ($\times8$)  & 31.32 & 23.15 & 2.18 \\ %
			\bottomrule
		\end{tabular}
	\end{center}
    \vspace{-10pt}
\end{table}

\begin{table}[!tb]
	\centering
	\caption{
		\textbf{Details of Heterogeneous Ensembles in~\cref{fig:homo_vs_hetero} and~\cref{tab:ensemble_type}}.
	}
	\label{tab:model_list}
	\setlength{\tabcolsep}{0.5em}
	\begin{tabular}{c|c}
		\toprule
		\multirow{3}{*}{\parbox{1.7cm}{\centering Weak models\\($69\unsim70\%$)}} & ResNet-18~\cite{he2016resnet},GoogleNet~\cite{szegedy2015going}, \\
		& VGG-11, VGG-13, VGG-11-BN, VGG-16~\cite{simonyan2014very}, \\
		& HRNet-W18~\cite{wang2020deep}, DLA~\cite{yu2018deep} \\
		\midrule
		  & ResNet-50~\cite{he2016resnet}, DenseNet-121~\cite{huang2017densenet}, \\
		Strong models & Inception-V3~\cite{szegedy2016rethinking}, VGG-19-BN~\cite{simonyan2014very}, \\
		($75\unsim76\%$) & RegNetY~\cite{radosavovic2020designing}, RepVGG-A2~\cite{ding2021repvgg}, \\
		& DPN-68~\cite{chen2016training}, DLA-X-60-C~\cite{yu2018deep} \\
		\bottomrule
	\end{tabular}
\end{table}

In this subsection, we study the design choices of \elodi.
Specifically, we first compare the effect of using homogeneous and heterogeneous ensembles as guidance.
Next, we empirically justify that the optimal architecture of the homogeneous ensemble in \elodi is the same as the model to be optimized. 
Finally, we empirically compare different choices of distillation functions.

\myparagraph{Homogeneous \versus heterogeneous ensembles}. 
We use a homogeneous ensemble in both \cref{sec:probe:landscape} and \ref{sec:method:elodi}.
However, in ensemble learning, members with strong diversity such as model architectures are usually favored for better generalization.
To study the potential advantage of homogeneous ensembles in \elodi over heterogeneous ones, we construct an \textbf{all-diff-weak} ensemble, which is composed of 8 different weak models with Top-1 Accuracy $=69\unsim70\%$ on ImageNet.
Similarly, we build an \textbf{all-diff-strong} ensemble, which is composed of 8 different strong models with Top-1 Accuracy $=75\unsim76\%$ on ImageNet,
The model list can be found in \cref{tab:model_list} and the model weights are adopted from \texttt{timm}~\cite{rw2019timm}.

In \cref{tab:ensemble_type}, we observe that using a homogeneous ensemble for guidance achieves comparable or slightly better results in both NFR and ER than heterogeneous (``all-different'') ensembles.
Some may argue that the confidence score might vary across different architectures.
To rule out potential issues on miscalibration, we also additionally calibrate ``all-different'' ensembles using temporal scaling~\cite{guo2017calibration} but observe no gain.
This suggests that strong diversity in a guiding ensemble may not lead to lower NFR.
Also, \elodi with homogeneous ensembles is easier to implement and extend in practice - a homogeneous ensemble requires only one architecture and needs neither re-calibrating prediction score nor balancing weights.

\myparagraph{Change of architecture for the guiding ensemble}. 
In \cref{tab:ensemble_type}, we find that training a new model guided by an ensemble with the old model's architecture has to trade ER for NFR reduction, which is not desired. 
This corroborates with the hypothesis in \cref{sec:probe:landscape} that models with different architectures have different representation landscapes and thus it is better to use an ensemble with the same architecture for guiding \elodi.
When a system has undergone multiple updates, always guiding \elodi with the new model's architecture also provides a clear guideline for practice.

\begin{table}[!tb]
	\centering
	\caption{
			\textbf{Distilling ensembles with different loss functions}.
			Considering the model update of ResNet-18$\rightarrow$ResNet-50, \elodi achieves lower NFR and ER than KD/FD.
	}
	\label{tab:ensemble_distill}
	\setlength{\tabcolsep}{0.5em}
	\begin{tabular}{l|c|c|c}
		\toprule
		\multirow{2}{*}{Method} & \multicolumn{2}{c|}{ER\lowbetter (\%)}   & NFR\lowbetter (\%) \\
		& RN-18\distill & RN-50\distill  & \\
		\cmidrule{1-4}
		\elodi & 30.95 & 23.10 & 2.11 \\
		Ensemble w/. KD$_{\tau=100}$ & 32.09 & 23.67 & 2.23 \\
        Ensemble w/. FD$_{\tau=100}$ & 32.19 & 23.97 & 2.16 \\
		Ensemble w/. FD$_{\tau=1}$  & 31.62 & 24.06 & 2.43 \\
		\bottomrule
	\end{tabular}
\end{table}

\myparagraph{Choices of distillation functions}.
We compare different choices of distillation loss functions in \cref{tab:ensemble_distill}.
We can see that \elodi outperforms FD~\cite{yan2021pct} (Ensemble w/. FD) or KD~\cite{hinton2015kd} (Ensemble w/. KD) in both ER and NFR.

\subsection{Exploratory Studies of Parameters in \elodi}
\label{sec:exp:elodi_more_ablations}

In this subsection, we provide more detailed ablation on the hyper-parameters in \elodi, namely the loss weight and size of the reference ensemble.
We end this subsection with a comparison between offline and online distillation.

\begin{figure}[!tb]
	\begin{center}
		\subfloat[{\small Varying loss weights.}]{
			\includegraphics[width=0.48\linewidth]{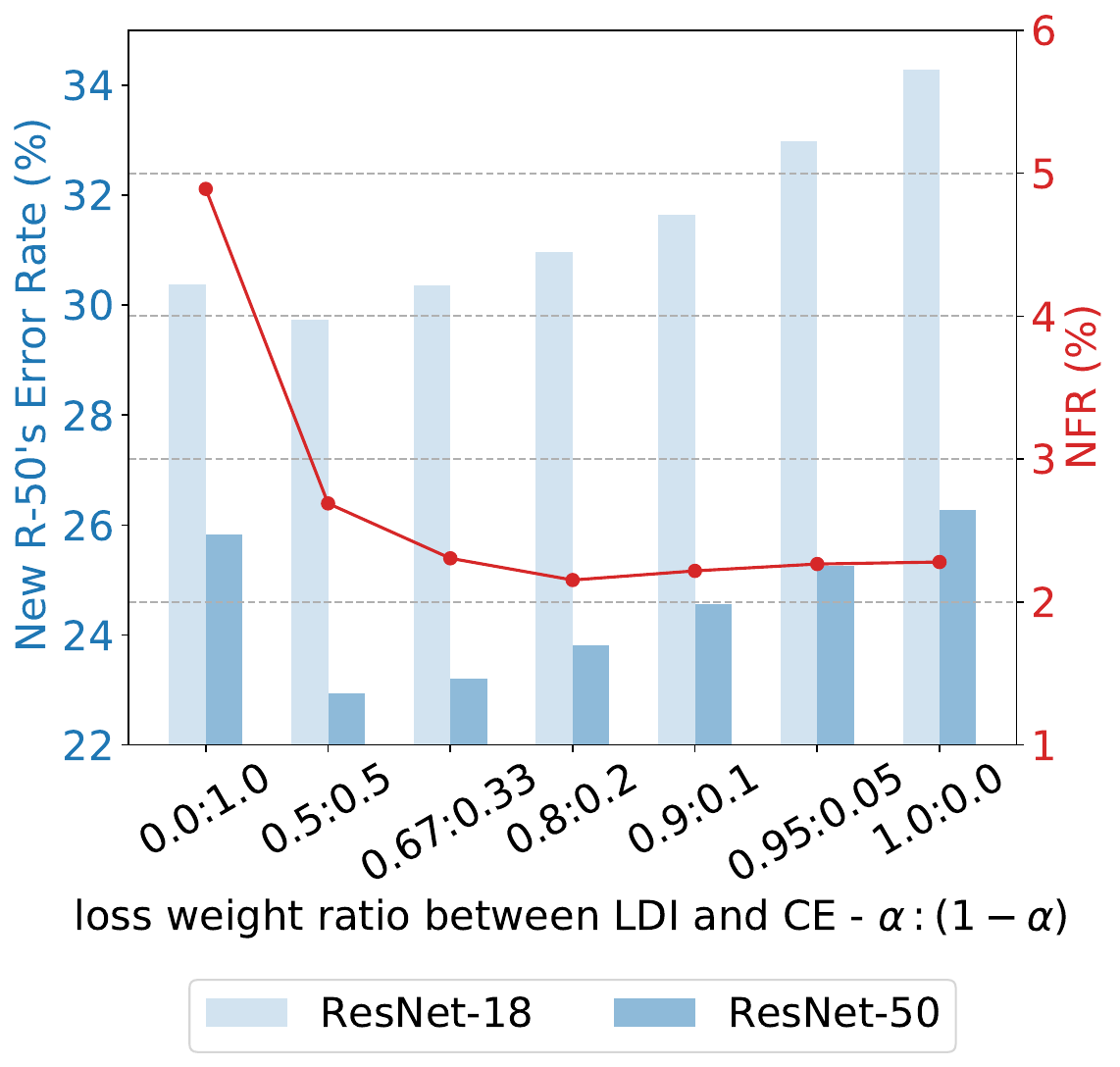}
			\label{fig:nfr_vs_loss_weight}
		}
		\subfloat[{\small Varying ensemble size.}]{
			\includegraphics[width=0.48\linewidth]{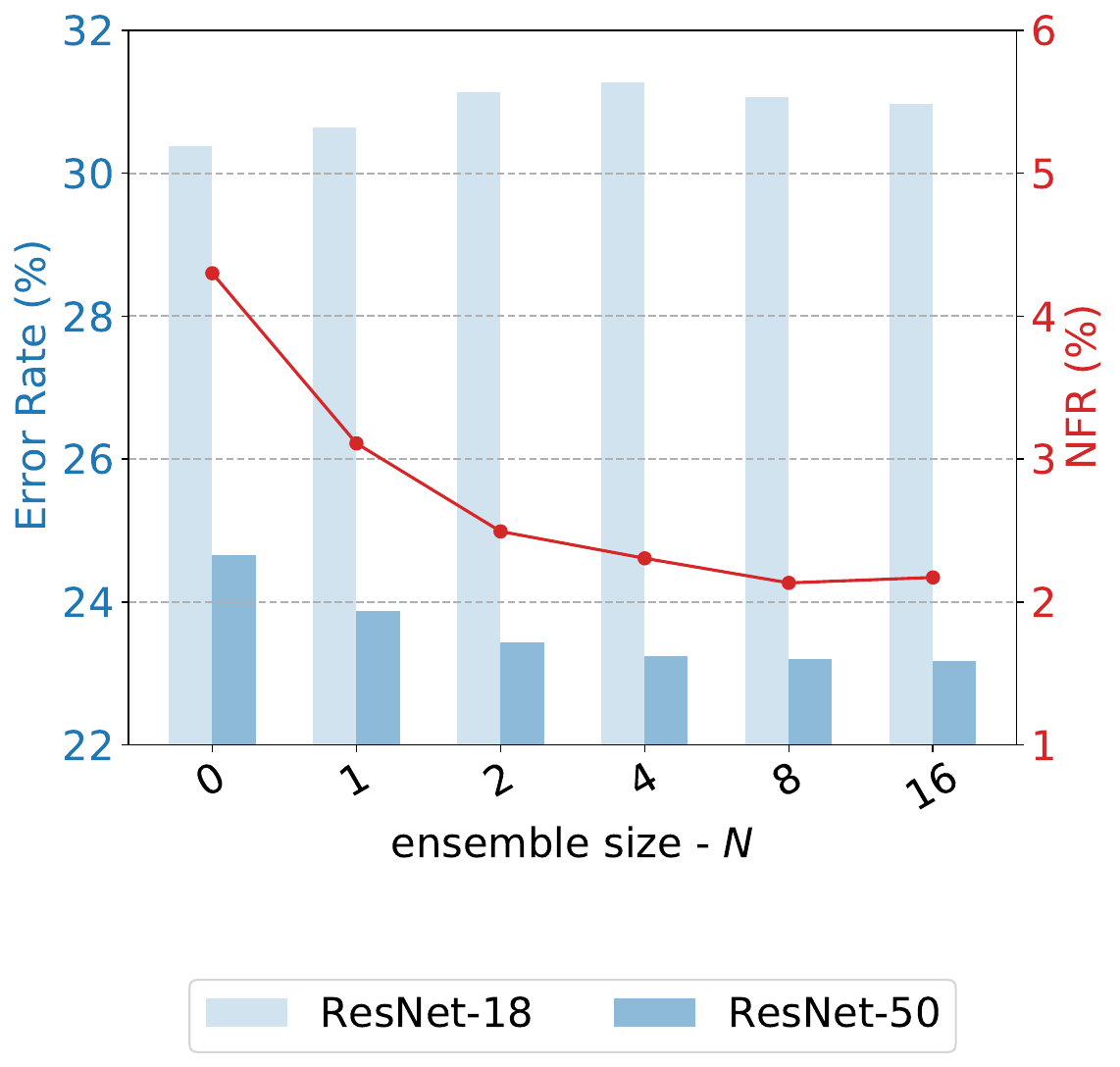}
			\label{fig:nfr_vs_ens_size}
		}
		\caption{
				\textbf{Ablating loss weights and ensemble sizes for \elodi on ImageNet.}
				ER of ResNet-18 (-50) is shown in the light (dark) blue bar plot while NFR is shown in the red curve.
		}%
		\label{fig:hyper_param}
	\end{center}
    \vspace{-10pt}
\end{figure}

\begin{table}[!tb]
	\centering
	\caption{
			\textbf{Comparison between offline and online distill on ImageNet.}
			Inferring teacher logits during training (online) achieves both lower ER and NFR compared to pre-extracting it (offline).}%
	\label{tab:offline_distillation}
	\begin{tabular}{l|c|c|c}
		\toprule
		\multirow{2}{*}{Method} & \multicolumn{2}{c|}{Error Rate\lowbetter (\%)}   & NFR\lowbetter (\%) \\
		\cmidrule{2-3}
		& RN-18\distill & RN-50\distill  &  \\
		\cmidrule{1-4}
		Offline  & 32.49 & 24.26 & 2.38 \\
		Online  & 30.97 & 23.81 & 2.15 \\
		\bottomrule
	\end{tabular}
\end{table}

\myparagraph{The effect of loss weight.}
We experiment with different loss weights $\alpha$ and summarize the results in \cref{fig:nfr_vs_loss_weight}.
$\alpha_\mathrm{\elodi}=0$ is equivalent to the no-treatment baseline where only the standard Cross-Entropy loss is applied.
$\alpha_\mathrm{\elodi}=1$ means that we drop the Cross-Entropy loss and distill logits only.
When $\alpha_\mathrm{LDI}$ increases from 0.5 to 1, the distilled model's ER first decreases and then increases for both models.
On the other hand, NFR consistently decreases and stays at around 2.2\%.
We find $\alpha_\mathrm{\elodi}=0.8$ achieves a good balance between the distilled model's ER and NFR. 
Therefore we use it by default for all \elodi experiments. 

\myparagraph{The size of reference ensemble.}
We study \elodi's effectiveness for reducing NFR by varying the ensemble size $m$ in \cref{fig:nfr_vs_ens_size}.
The case of $m=0$ is the no-treatment baseline.
The case of $m=1$ can be viewed as self-distillation~\cite{zhang2019selfdistill} except that the new model's weight is re-initialized.
We can see that NFR consistently decreases from 4.30\% to 2.15\% when the ensemble size increases from 1 to 8.

\myparagraph{Online~\versus offline distillation.} 
In \elodi, the ensemble's logits can be either inferred during training (online) or pre-extracted before training (offline).
In~\cref{tab:offline_distillation}, we find that offline distillation is less effective in reducing NFR and ER.
A similar observation is also reported in~\cite{beyer2022knowledge}.
Therefore we use the online approach in all experiments.

\section{Conclusion and Discussion}

\noindent Our experiments show that \elodi performs positive congruent training by reducing negative flips with large logit displacement and reducing the variance of logits from the ensemble estimates. 
This behavior can be transferred to single models through the \elodi method can benefit updates with single models. 

As discussed in~\cref{sec:probe:landscape} and later observed in experiments, the difference in representation landscape could still lead to non-zero NFR in the updates even with \elodi, which requires future works on in-depth characterizing of representation landscape change between model architectures. 
Another limitation of \elodi is that the training cost is still higher than the normal training process of a classification model update, due to the additional training of the ensemble and online inference of the ensemble logits, calling for further efficiency improvement.

\ifCLASSOPTIONcaptionsoff
  \newpage
\fi

\bibliography{ref}

\begin{thebibliography}{10}
\providecommand{\url}[1]{#1}
\csname url@samestyle\endcsname
\providecommand{\newblock}{\relax}
\providecommand{\bibinfo}[2]{#2}
\providecommand{\BIBentrySTDinterwordspacing}{\spaceskip=0pt\relax}
\providecommand{\BIBentryALTinterwordstretchfactor}{4}
\providecommand{\BIBentryALTinterwordspacing}{\spaceskip=\fontdimen2\font plus
\BIBentryALTinterwordstretchfactor\fontdimen3\font minus
  \fontdimen4\font\relax}
\providecommand{\BIBforeignlanguage}[2]{{%
\expandafter\ifx\csname l@#1\endcsname\relax
\typeout{** WARNING: IEEEtran.bst: No hyphenation pattern has been}%
\typeout{** loaded for the language `#1'. Using the pattern for}%
\typeout{** the default language instead.}%
\else
\language=\csname l@#1\endcsname
\fi
#2}}
\providecommand{\BIBdecl}{\relax}
\BIBdecl

\bibitem{yan2021pct}
S.~Yan, Y.~Xiong, K.~Kundu, S.~Yang, S.~Deng, M.~Wang, W.~Xia, and S.~Soatto,
  ``Positive-congruent training: Towards regression-free model updates,'' in
  \emph{CVPR}, 2021.

\bibitem{bansal2019updates}
G.~Bansal, B.~Nushi, E.~Kamar, D.~S. Weld, W.~S. Lasecki, and E.~Horvitz,
  ``Updates in human-ai teams: Understanding and addressing the
  performance/compatibility tradeoff,'' in \emph{AAAI}, 2019.

\bibitem{xie2021regression}
Y.~Xie, Y.-a. Lai, Y.~Xiong, Y.~Zhang, and S.~Soatto, ``Regression bugs are in
  your model! measuring, reducing and analyzing regressions in nlp model
  updates,'' in \emph{ACL}, 2021.

\bibitem{hinton2015kd}
G.~Hinton, O.~Vinyals, and J.~Dean, ``Distilling the knowledge in a neural
  network,'' \emph{arXiv preprint arXiv:1503.02531}, 2015.

\bibitem{lakshminarayanan2017deepensemble}
B.~Lakshminarayanan, A.~Pritzel, and C.~Blundell, ``Simple and scalable
  predictive uncertainty estimation using deep ensembles,'' in \emph{NeurIPS},
  2017.

\bibitem{toneva2019empirical}
M.~Toneva, A.~Sordoni, R.~T.~d. Combes, A.~Trischler, Y.~Bengio, and G.~J.
  Gordon, ``An empirical study of example forgetting during deep neural network
  learning,'' in \emph{ICLR}, 2019.

\bibitem{shen2020bct}
Y.~Shen, Y.~Xiong, W.~Xia, and S.~Soatto, ``Towards backward-compatible
  representation learning,'' in \emph{CVPR}, 2020.

\bibitem{srivastava2020empirical}
M.~Srivastava, B.~Nushi, E.~Kamar, S.~Shah, and E.~Horvitz, ``An empirical
  analysis of backward compatibility in machine learning systems,'' in
  \emph{KDD}, 2020.

\bibitem{jiang2022churn}
H.~Jiang, H.~Narasimhan, D.~Bahri, A.~Cotter, and A.~Rostamizadeh, ``Churn
  reduction via distillation,'' in \emph{ICLR}, 2022.

\bibitem{trauble2021backward}
F.~Tr{\"a}uble, J.~von K{\"u}gelgen, M.~Kleindessner, F.~Locatello,
  B.~Sch{\"o}lkopf, and P.~Gehler, ``Backward-compatible prediction updates: A
  probabilistic approach,'' in \emph{NeurIPS}, 2021.

\bibitem{breiman1996bagging}
L.~Breiman, ``Bagging predictors,'' \emph{Machine learning}, vol.~24, no.~2,
  pp. 123--140, 1996.

\bibitem{freund1997adaboost}
Y.~Freund and R.~E. Schapire, ``A decision-theoretic generalization of on-line
  learning and an application to boosting,'' \emph{Journal of computer and
  system sciences}, vol.~55, no.~1, pp. 119--139, 1997.

\bibitem{breiman2001randomforest}
L.~Breiman, ``Random forests,'' \emph{Machine learning}, vol.~45, no.~1, pp.
  5--32, 2001.

\bibitem{bartlett1998boosting}
P.~Bartlett, Y.~Freund, W.~S. Lee, and R.~E. Schapire, ``Boosting the margin: A
  new explanation for the effectiveness of voting methods,'' \emph{The annals
  of statistics}, vol.~26, no.~5, pp. 1651--1686, 1998.

\bibitem{allen2023towards}
Z.~Allen-Zhu and Y.~Li, ``Towards understanding ensemble, knowledge
  distillation and self-distillation in deep learning,'' in \emph{ICLR}, 2023.

\bibitem{hoeting1999bayesian}
\BIBentryALTinterwordspacing
J.~A. Hoeting, D.~Madigan, A.~E. Raftery, and C.~T. Volinsky, ``Bayesian model
  averaging: a tutorial,'' \emph{Statistical Science}, vol.~14, no.~4, pp. 382
  -- 417, 1999. [Online]. Available:
  \url{https://doi.org/10.1214/ss/1009212519}
\BIBentrySTDinterwordspacing

\bibitem{fragoso2018bayesian}
T.~M. Fragoso, W.~Bertoli, and F.~Louzada, ``Bayesian model averaging: A
  systematic review and conceptual classification,'' \emph{International
  Statistical Review}, vol.~86, no.~1, pp. 1--28, 2018.

\bibitem{huang2017snapshot}
G.~Huang, Y.~Li, G.~Pleiss, Z.~Liu, J.~E. Hopcroft, and K.~Q. Weinberger,
  ``Snapshot ensembles: Train 1, get m for free,'' in \emph{ICLR}, 2017.

\bibitem{garipov2018loss}
T.~Garipov, P.~Izmailov, D.~Podoprikhin, D.~P. Vetrov, and A.~G. Wilson, ``Loss
  surfaces, mode connectivity, and fast ensembling of dnns,'' in
  \emph{NeurIPS}, 2018.

\bibitem{srivastava2014dropout}
N.~Srivastava, G.~Hinton, A.~Krizhevsky, I.~Sutskever, and R.~Salakhutdinov,
  ``Dropout: a simple way to prevent neural networks from overfitting,''
  \emph{JMLR}, 2014.

\bibitem{gal2016dropout}
Y.~Gal and Z.~Ghahramani, ``Dropout as a bayesian approximation: Representing
  model uncertainty in deep learning,'' in \emph{ICML}, 2016.

\bibitem{larsson2017fractalnet}
G.~Larsson, M.~Maire, and G.~Shakhnarovich, ``Fractalnet: Ultra-deep neural
  networks without residuals,'' in \emph{ICLR}, 2017.

\bibitem{huang2016deep}
G.~Huang, Y.~Sun, Z.~Liu, D.~Sedra, and K.~Q. Weinberger, ``Deep networks with
  stochastic depth,'' in \emph{EECV}, 2016.

\bibitem{wen2020batchensemble}
Y.~Wen, D.~Tran, and J.~Ba, ``Batchensemble: an alternative approach to
  efficient ensemble and lifelong learning,'' in \emph{ICLR}, 2020.

\bibitem{havasi2021mimo}
M.~Havasi, R.~Jenatton, S.~Fort, J.~Z. Liu, J.~Snoek, B.~Lakshminarayanan,
  A.~M. Dai, and D.~Tran, ``Training independent subnetworks for robust
  prediction,'' in \emph{ICLR}, 2021.

\bibitem{zhang2019selfdistill}
L.~Zhang, J.~Song, A.~Gao, J.~Chen, C.~Bao, and K.~Ma, ``Be your own teacher:
  Improve the performance of convolutional neural networks via self
  distillation,'' in \emph{ICCV}, 2019.

\bibitem{yuan2020revisiting}
L.~Yuan, F.~E. Tay, G.~Li, T.~Wang, and J.~Feng, ``Revisiting knowledge
  distillation via label smoothing regularization,'' in \emph{CVPR}, 2020.

\bibitem{reich2020ensemble}
S.~Reich, D.~Mueller, and N.~Andrews, ``Ensemble distillation for structured
  prediction: Calibrated, accurate, fast-choose three,'' in \emph{EMNLP}, 2020.

\bibitem{fukuda2017efficient}
T.~Fukuda, M.~Suzuki, G.~Kurata, S.~Thomas, J.~Cui, and B.~Ramabhadran,
  ``Efficient knowledge distillation from an ensemble of teachers.'' in
  \emph{Interspeech}, 2017.

\bibitem{asif2020ensemble}
U.~Asif, J.~Tang, and S.~Harrer, ``Ensemble knowledge distillation for learning
  improved and efficient networks,'' in \emph{ECAI}, 2020.

\bibitem{malinin2020ensemble}
A.~Malinin, B.~Mlodozeniec, and M.~Gales, ``Ensemble distribution
  distillation,'' in \emph{ICLR}, 2020.

\bibitem{lin2020ensemble}
T.~Lin, L.~Kong, S.~U. Stich, and M.~Jaggi, ``Ensemble distillation for robust
  model fusion in federated learning,'' in \emph{NeurIPS}, 2020.

\bibitem{kuncheva2003measures}
L.~I. Kuncheva and C.~J. Whitaker, ``Measures of diversity in classifier
  ensembles and their relationship with the ensemble accuracy,'' \emph{Machine
  learning}, vol.~51, no.~2, pp. 181--207, 2003.

\bibitem{gontijo2022no}
R.~Gontijo-Lopes, Y.~Dauphin, and E.~D. Cubuk, ``No one representation to rule
  them all: Overlapping features of training methods,'' in \emph{ICLR}, 2022.

\bibitem{wortsman2022modelsoups}
M.~Wortsman, G.~Ilharco, S.~Y. Gadre, R.~Roelofs, R.~Gontijo-Lopes, A.~S.
  Morcos, H.~Namkoong, A.~Farhadi, Y.~Carmon \emph{et~al.}, ``Model soups:
  averaging weights of multiple fine-tuned models improves accuracy without
  increasing inference time,'' in \emph{ICML}, 2022.

\bibitem{rvavceva1962domains}
E.~Rva{\v{c}}eva, ``On domains of attraction of multi-dimensional
  distributions,'' \emph{Selected Translations in Mathematical Statistics and
  Probability}, vol.~2, pp. 183--205, 1962.

\bibitem{ferguson2017course}
T.~S. Ferguson, \emph{A course in large sample theory}.\hskip 1em plus 0.5em
  minus 0.4em\relax Routledge, 2017.

\bibitem{mathai1992quadratic}
A.~M. Mathai and S.~B. Provost, \emph{Quadratic forms in random variables:
  theory and applications}.\hskip 1em plus 0.5em minus 0.4em\relax Dekker,
  1992.

\bibitem{das2021method}
A.~Das and W.~S. Geisler, ``A method to integrate and classify normal
  distributions,'' \emph{Journal of Vision}, vol.~21, no.~10, 2021.

\bibitem{deng2009imagenet}
J.~Deng, W.~Dong, R.~Socher, L.-J. Li, K.~Li, and L.~Fei-Fei, ``Imagenet: A
  large-scale hierarchical image database,'' in \emph{CVPR}, 2009.

\bibitem{seguin2021understanding}
L.~Seguin, A.~Ndirango, N.~Mishra, S.~Chung, and T.~Lee, ``Understanding the
  logit distributions of adversarially-trained deep neural networks,''
  \emph{arXiv preprint arXiv:2108.12001}, 2021.

\bibitem{parzen1962kde}
E.~Parzen, ``On estimation of a probability density function and mode,''
  \emph{The annals of mathematical statistics}, vol.~33, no.~3, 1962.

\bibitem{glorot2010xavier}
X.~Glorot and Y.~Bengio, ``Understanding the difficulty of training deep
  feedforward neural networks,'' in \emph{AISTATS}, 2010.

\bibitem{he2015kaiming}
K.~He, X.~Zhang, S.~Ren, and J.~Sun, ``Delving deep into rectifiers: Surpassing
  human-level performance on imagenet classification,'' in \emph{ICCV}, 2015.

\bibitem{chizat2019lazy}
L.~Chizat, E.~Oyallon, and F.~Bach, ``On lazy training in differentiable
  programming,'' in \emph{NeurIPS}, 2019.

\bibitem{romero2015fitnet}
A.~Romero, N.~Ballas, S.~E. Kahou, A.~Chassang, C.~Gatta, and Y.~Bengio,
  ``Fitnets: Hints for thin deep nets,'' in \emph{ICLR}, 2015.

\bibitem{zagoruyko2017paying}
S.~Zagoruyko and N.~Komodakis, ``Paying more attention to attention: Improving
  the performance of convolutional neural networks via attention transfer,'' in
  \emph{ICLR}, 2017.

\bibitem{yim2017gift}
J.~Yim, D.~Joo, J.~Bae, and J.~Kim, ``A gift from knowledge distillation: Fast
  optimization, network minimization and transfer learning,'' in \emph{CVPR},
  2017.

\bibitem{an2020partialfc}
X.~An, X.~Zhu, Y.~Xiao, L.~Wu, M.~Zhang, Y.~Gao, B.~Qin, D.~Zhang, and F.~Ying,
  ``Partial fc: Training 10 million identities on a single machine,'' in
  \emph{Arxiv 2010.05222}, 2020.

\bibitem{van2018inaturalist}
G.~Van~Horn, O.~Mac~Aodha, Y.~Song, Y.~Cui, C.~Sun, A.~Shepard, H.~Adam,
  P.~Perona, and S.~Belongie, ``The {iNaturalist} species classification and
  detection dataset,'' in \emph{CVPR}, 2018.

\bibitem{russakovsky2015imagenet}
O.~Russakovsky, J.~Deng, H.~Su, J.~Krause, S.~Satheesh, S.~Ma, Z.~Huang,
  A.~Karpathy, A.~Khosla, M.~Bernstein \emph{et~al.}, ``Imagenet large scale
  visual recognition challenge,'' \emph{IJCV}, 2015.

\bibitem{zhang2015character}
X.~Zhang, J.~Zhao, and Y.~LeCun, ``Character-level convolutional networks for
  text classification,'' in \emph{NeurIPS}, 2015.

\bibitem{chen2016training}
T.~Chen, B.~Xu, C.~Zhang, and C.~Guestrin, ``Training deep nets with sublinear
  memory cost,'' \emph{arXiv preprint arXiv:1604.06174}, 2016.

\bibitem{goyal2017accurate}
P.~Goyal, P.~Doll{\'a}r, R.~Girshick, P.~Noordhuis, L.~Wesolowski, A.~Kyrola,
  A.~Tulloch, Y.~Jia, and K.~He, ``Accurate, large minibatch sgd: Training
  imagenet in 1 hour,'' \emph{arXiv preprint arXiv:1706.02677}, 2017.

\bibitem{zhang2022ract}
B.~Zhang, Y.~Ge, Y.~Shen, Y.~Li, C.~Yuan, X.~Xu, Y.~Wang, and Y.~Shan,
  ``Hot-refresh model upgrades with regression-free compatible training in
  image retrieval,'' in \emph{ICLR}, 2022.

\bibitem{devlin2018bert}
J.~Devlin, M.-W. Chang, K.~Lee, and K.~Toutanova, ``Bert: Pre-training of deep
  bidirectional transformers for language understanding,'' in \emph{NAACL},
  2018.

\bibitem{sun2019fine}
C.~Sun, X.~Qiu, Y.~Xu, and X.~Huang, ``How to fine-tune {BERT} for text
  classification?'' in \emph{CCL}, 2019.

\bibitem{guo2017calibration}
C.~Guo, G.~Pleiss, Y.~Sun, and K.~Q. Weinberger, ``On calibration of modern
  neural networks,'' in \emph{ICML}, 2017.

\bibitem{he2016resnet}
K.~He, X.~Zhang, S.~Ren, and J.~Sun, ``Deep residual learning for image
  recognition,'' in \emph{CVPR}, 2016.

\bibitem{szegedy2015going}
C.~Szegedy, W.~Liu, Y.~Jia, P.~Sermanet, S.~Reed, D.~Anguelov, D.~Erhan,
  V.~Vanhoucke, and A.~Rabinovich, ``Going deeper with convolutions,'' in
  \emph{CVPR}, 2015.

\bibitem{simonyan2014very}
K.~Simonyan and A.~Zisserman, ``Very deep convolutional networks for
  large-scale image recognition,'' \emph{arXiv preprint arXiv:1409.1556}, 2014.

\bibitem{wang2020deep}
J.~Wang, K.~Sun, T.~Cheng, B.~Jiang, C.~Deng, Y.~Zhao, D.~Liu, Y.~Mu, M.~Tan,
  X.~Wang \emph{et~al.}, ``Deep high-resolution representation learning for
  visual recognition,'' \emph{TPAMI}, 2020.

\bibitem{yu2018deep}
F.~Yu, D.~Wang, E.~Shelhamer, and T.~Darrell, ``Deep layer aggregation,'' in
  \emph{CVPR}, 2018.

\bibitem{huang2017densenet}
G.~Huang, Z.~Liu, L.~Van Der~Maaten, and K.~Q. Weinberger, ``Densely connected
  convolutional networks,'' in \emph{CVPR}, 2017.

\bibitem{szegedy2016rethinking}
C.~Szegedy, V.~Vanhoucke, S.~Ioffe, J.~Shlens, and Z.~Wojna, ``Rethinking the
  inception architecture for computer vision,'' in \emph{CVPR}, 2016.

\bibitem{radosavovic2020designing}
I.~Radosavovic, R.~P. Kosaraju, R.~Girshick, K.~He, and P.~Doll{\'a}r,
  ``Designing network design spaces,'' in \emph{CVPR}, 2020.

\bibitem{ding2021repvgg}
X.~Ding, X.~Zhang, N.~Ma, J.~Han, G.~Ding, and J.~Sun, ``Rep{VGG}: Making
  vgg-style convnets great again,'' in \emph{CVPR}, 2021.

\bibitem{rw2019timm}
R.~Wightman, ``Pytorch image models,''
  \url{https://github.com/rwightman/pytorch-image-models}, 2019.

\bibitem{beyer2022knowledge}
L.~Beyer, X.~Zhai, A.~Royer, L.~Markeeva, R.~Anil, and A.~Kolesnikov,
  ``Knowledge distillation: A good teacher is patient and consistent,'' in
  \emph{CVPR}, 2022.

\end{thebibliography}
\bibliographystyle{IEEEtran}

\vspace{-20pt}

\begin{IEEEbiography}[{\includegraphics[width=1in,height=1.25in,clip,keepaspectratio]{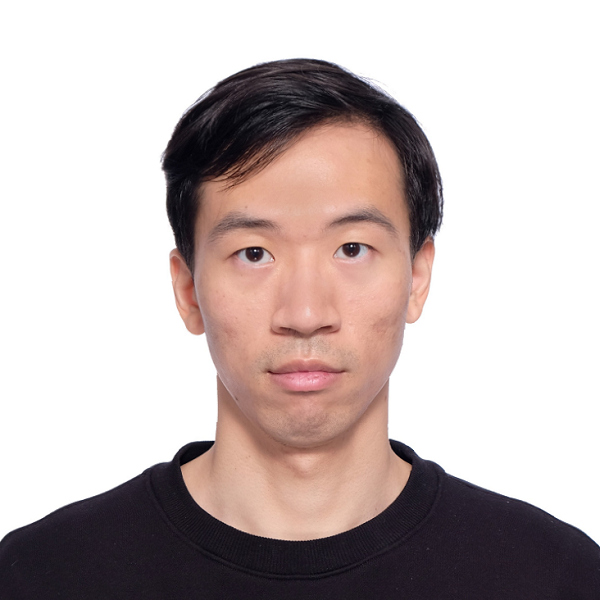}}]{Yue Zhao}
is currently a PhD student at the University of Texas at Austin, supervised by Prof. Philipp Kr\"ahenb\"uhl.
He obtained his MPhil's degree from the Chinese University of Hong Kong, supervised by Prof. Dahua Lin. 
Previously, he got his BE's degree from Tsinghua University.
His research interests are in computer vision, particularly video understanding.
He is a recipient of the NVIDIA Graduate Fellowship.
\end{IEEEbiography}

\vspace{-20pt}

\begin{IEEEbiography}[{\includegraphics[width=1in,height=1.25in,clip,keepaspectratio]{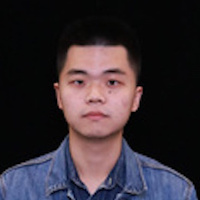}}]{Yantao Shen}
is an applied scientist at AWS AI.
Yantao Shen received a BE degree from Northeastern University, Shenyang, China, in 2015, and a PhD degree in electronic
engineering from the Chinese University of Hong Kong, in 2020.
\end{IEEEbiography}

\vspace{-20pt}

\begin{IEEEbiography}[{\includegraphics[width=1in,height=1.25in,clip,keepaspectratio]{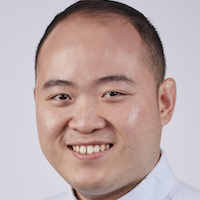}}]{Yuanjun Xiong}
is a principal applied scientist at Amazon Rekognition.
Before joining Amazon, he was a postdoctoral fellow in the Department of Information Engineering, at CUHK.
Yuanjun Xiong received a BS degree from Tsinghua University, Beijing, China, in 2012, and a PhD degree in information engineering from the Chinese University of Hong Kong, in 2016.
His research interests include computer vision, image understanding, and video content analysis.
\end{IEEEbiography}

\vspace{-20pt}

\begin{IEEEbiography}[{\includegraphics[width=1in,height=1.25in,clip,keepaspectratio]{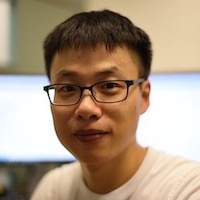}}]{Shuo Yang}
is currently a senior applied scientist at AWS-AI Computer Vision.
He received his Ph.D. from the Department of Information Engineering, CUHK in July 2017.
He is supervised by Prof. Xiaoou Tang and Prof. Chen Change Loy.
His research interests include face/object detection and object recognition.
Shuo graduated from the School of Software, Wuhan University with a B.Eng. in 2013.
\end{IEEEbiography}

\vspace{-15pt}

\begin{IEEEbiography}[{\includegraphics[width=1in,height=1.25in,clip,keepaspectratio]{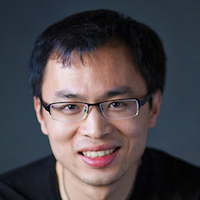}}]{Wei Xia}
is a Principal Scientist at AWS AI, leading the research for AWS ReKognition.
Previously he was the Chief scientist at Orbeus Inc, an AI startup acquired by Amazon in Sep 2015.
Wei obtained his Phd in computer vision and machine learning from National University of Singapore and performed as a visiting researcher at Panasonic and Lund University, respectively.
He has published around 40 papers and filed 30+ patents.
He also served as a reviewer and workshop organizer in top-tiered conferences and journals, like ICCV/CVPR/ECCV, etc.
During his years in AWS, he and his team started a new research area in model compatibility.
He has won the winner awards of Pascal VOC classification and segmentation competitions in 2012, runner-up award in ILSVRC challenge 2013 and winner award of ILSVRC detection challenge in 2014.
\end{IEEEbiography}

\vspace{-15pt}

\begin{IEEEbiography}[{\includegraphics[width=1in,height=1.25in,clip,keepaspectratio]{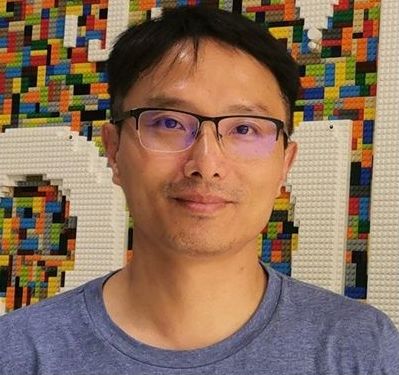}}]{Zhuowen Tu}
is a full professor of Cognitive Science and also affiliated with the Department of Computer Science and Engineering, University of California San Diego.
Before joining UCSD in 2013 as an assistant professor, he was a faculty member at UCLA.
Between 2011 and 2013, he took a leave to work at Microsoft Research Asia.
He received his Ph.D. from the Ohio State University and his M.E. from Tsinghua University.
He is a recipient of the David Marr Prize Award 2003 and a recipient of the David Marr Prize Honorable Mention Award 2015.
He is a Fellow of the IEEE.
\end{IEEEbiography}

\vspace{-15pt}

\begin{IEEEbiography}[{\includegraphics[width=1in,height=1.25in,clip,keepaspectratio]{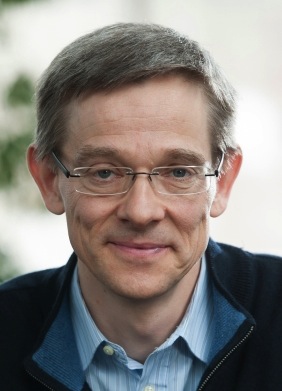}}]{Bernt Schiele}
has been Max Planck Director at MPI for Informatics and Professor at Saarland University since 2010.
He studied computer science at the University of Karlsruhe, Germany.
He worked on his master thesis in the field of robotics in Grenoble, France, where he also obtained the ``diplome d\'etudes approfondies d\'informatique''.
In 1994 he worked in the field of multi-modal human-computer interfaces at Carnegie Mellon University, Pittsburgh, PA,
USA in the group of Alex Waibel.
In 1997 he obtained his PhD from INP Grenoble, France under the supervision of Prof. James L. Crowley in the field of computer vision.
The title of his thesis was ``Object Recognition using Multidimensional Receptive Field Histograms''.
Between 1997 and 2000 he was postdoctoral associate and Visiting Assistant Professor with the group of Prof. Alex Pentland at the Media Laboratory of the Massachusetts Institute of Technology, Cambridge, MA, USA.
From 1999 until 2004 he was Assistant Professor at the Swiss Federal Institute of Technology in Zurich (ETH Zurich).
Between 2004 and 2010 he was Full Professor at the computer science department of TU Darmstadt.
\end{IEEEbiography}

\vspace{-15pt}

\begin{IEEEbiography}[{\includegraphics[width=1in,height=1.25in,clip,keepaspectratio]{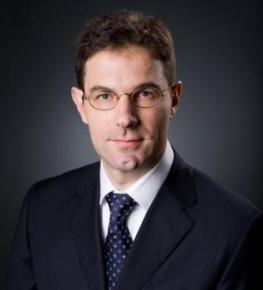}}]{Stefano Soatto}
is professor of computer science at the University of California, Los Angeles (UCLA), where he is also professor of electrical engineering and founding director of the UCLA Vision Lab.
He is also a VP in AWS AI.
He was named Fellow of the Institute of Electrical and Electronics Engineers (IEEE) in 2013 for contributions to dynamic visual processes.
He received the David Marr Prize in Computer Vision in 1999.
\end{IEEEbiography}

\clearpage

\appendices

\section{Visualization on More Data Points}
\label{appendix:visualization}

\begin{figure*}[t]
	\begin{center}
		\subfloat[{\small \texttt{val\_00009585}}.]{
		\includegraphics[width=0.21\linewidth]{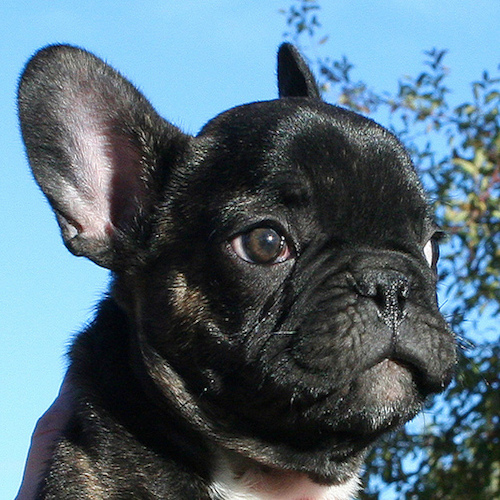}
		\label{supp_fig:more_data_example_337}
	}
		\subfloat[{\small \texttt{val\_00015098}}.]{
		\includegraphics[width=0.21\linewidth]{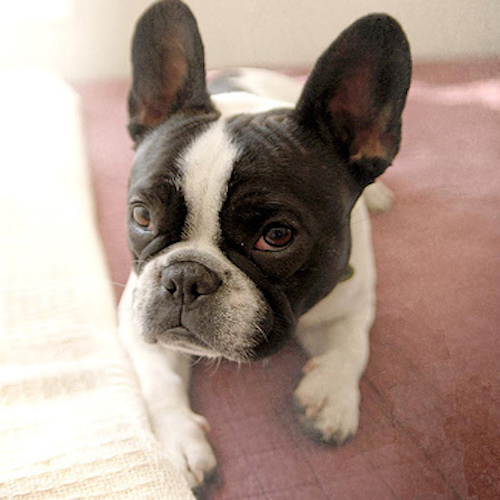}
		\label{supp_fig:more_data_example_399}
	}
		\subfloat[{\small \texttt{val\_00034619}}]{
		\includegraphics[width=0.21\linewidth]{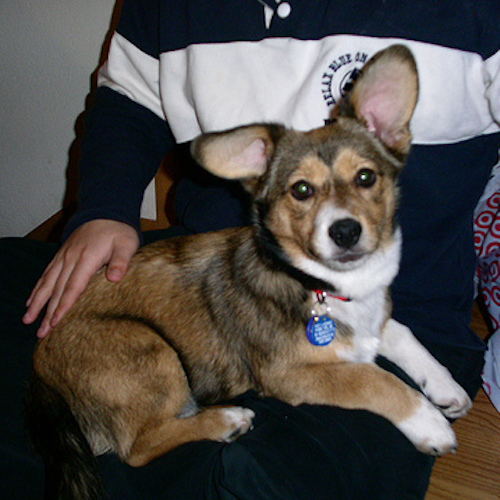}
		\label{supp_fig:more_data_example_217}
	}
		\subfloat[{\small \texttt{val\_00014560}}]{
		\includegraphics[width=0.21\linewidth]{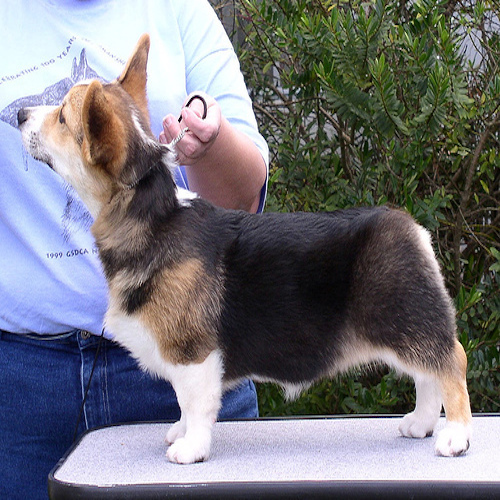}
		\label{supp_fig:more_data_example_431}
	}
		\caption{
				\textbf{Four example images for visualization}.
				The two classes we select here are ``French bulldog'' (n02108915) and ``Welsh Corgi'' (n02113023).
		}
		\label{supp_fig:more_data_example}
	\end{center}
    \vskip -0.1 in
\end{figure*}

\begin{figure*}[t]
	\begin{center}
		\subfloat[{\small \texttt{val\_00009585}.}]{
		\includegraphics[width=0.21\linewidth]{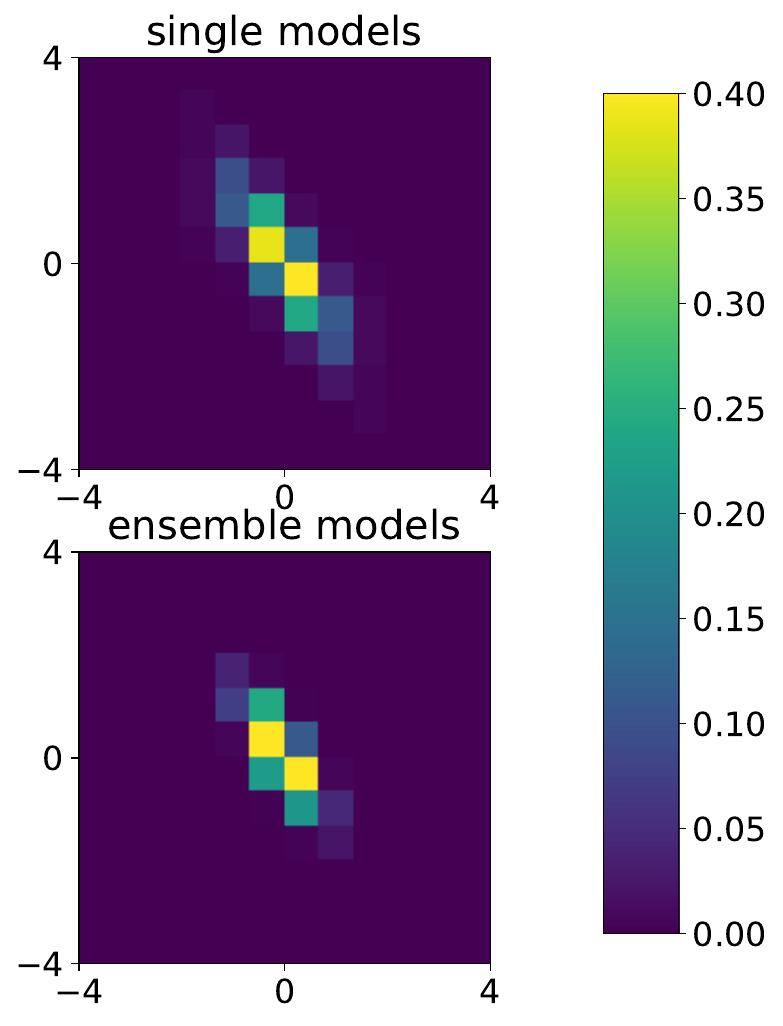}
		\label{supp_fig:hist2d_clt_337}
	}
		\subfloat[{\small \texttt{val\_00015098}.}]{
		\includegraphics[width=0.21\linewidth]{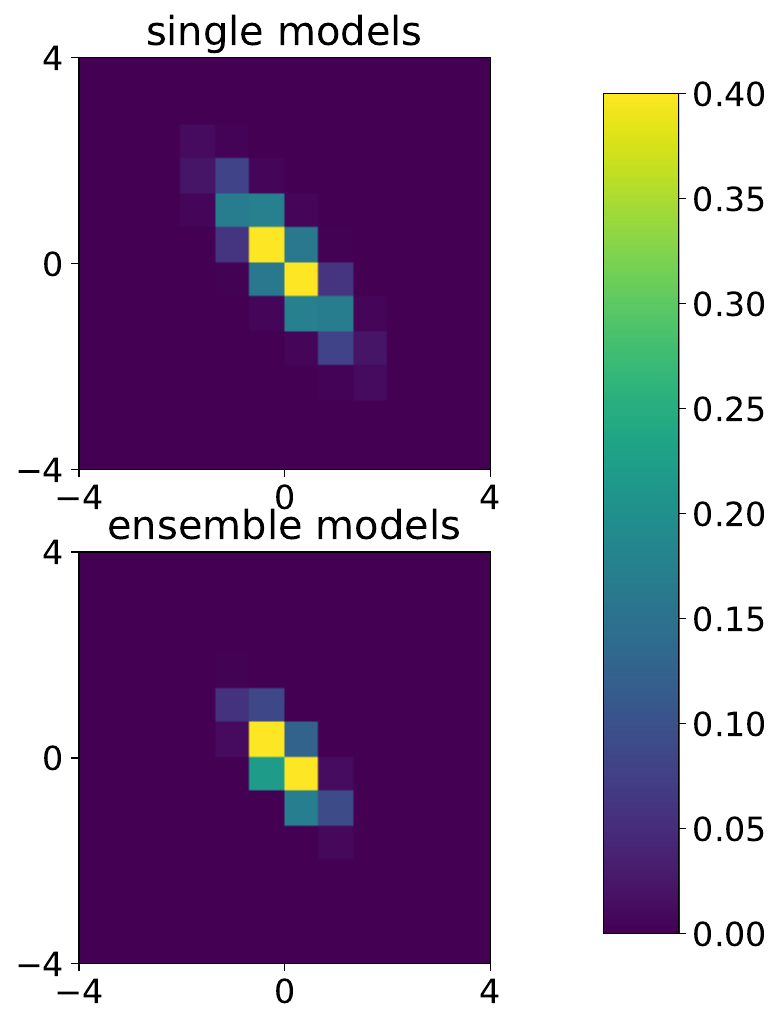}
		\label{supp_fig:hist2d_clt_399}
	}
		\subfloat[{\small \texttt{val\_00034619}.}]{
		\includegraphics[width=0.21\linewidth]{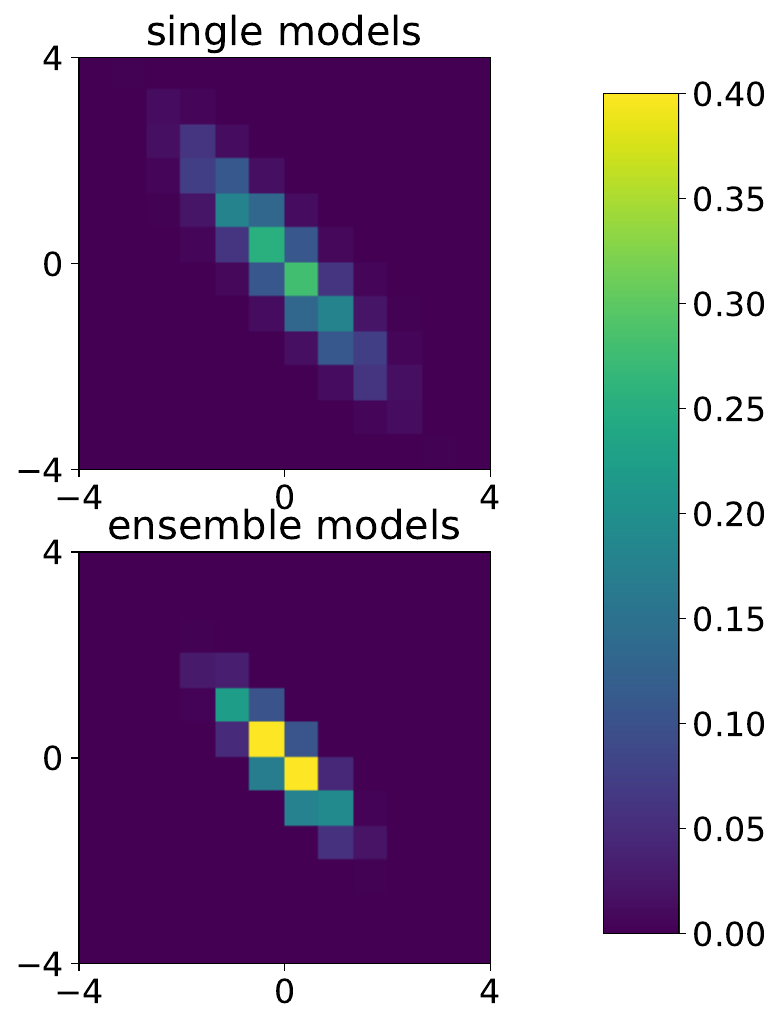}
		\label{supp_fig:hist2d_clt_217}
	}
		\subfloat[{\small \texttt{val\_00014560}.}]{
		\includegraphics[width=0.21\linewidth]{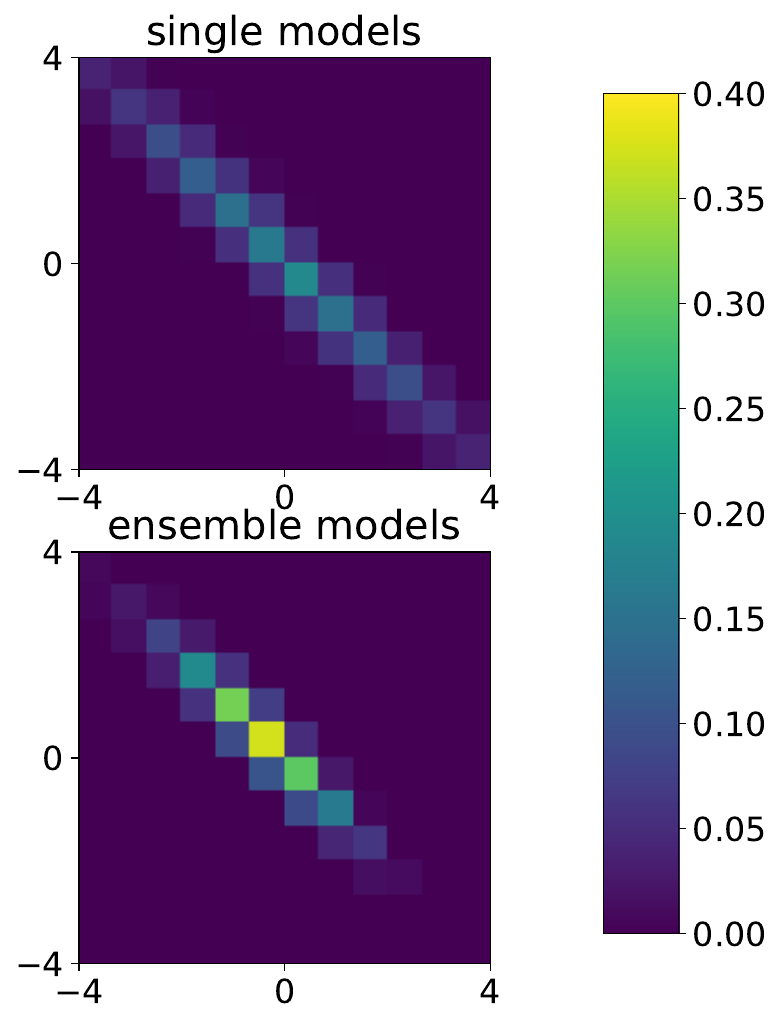}
		\label{supp_fig:hist2d_clt_431}
	}
	\caption{
		Estimated probability mass function (PMF) of logit displacement between two single models or ensembles.
		The $x,y$-axes denote the two classes' logit displacement.
		The heatmap value denotes the estimated probability density.
		The ensemble's co-variance is significantly smaller than the single model.
		The figure is best viewed in color.}
		\label{supp_fig:hist2d_clt_more_data}
	\end{center}
    \vskip -0.1 in
\end{figure*}

As mentioned in \cref{sec:prob:toy_example} and \cref{sec:probe:val}, we provide more examples to verify our hypothesis.
We select four images of two classes from ImageNet~\cite{deng2009imagenet}, which are illustrated in \cref{supp_fig:more_data_example}, as input data. With these input images, the estimated probability mass function (PMF) of logit displacement between two single models and two ensembles are shown in \cref{supp_fig:hist2d_clt_more_data}. We can observe that the logit displacements are reduced with ensembles, which verifies our hypothesis that output logit vectors are actually independent and identically distributed (\iid) random variables and with multi-dimensional central limit theorem (CLT), their sum is a normal distribution (\cref{eq:clt}).

\begin{figure*}[t]
	\begin{minipage}{\textwidth}
		\centering
		\subfloat[{\small \texttt{val\_00009585}.}]{
			\includegraphics[width=0.24\linewidth]{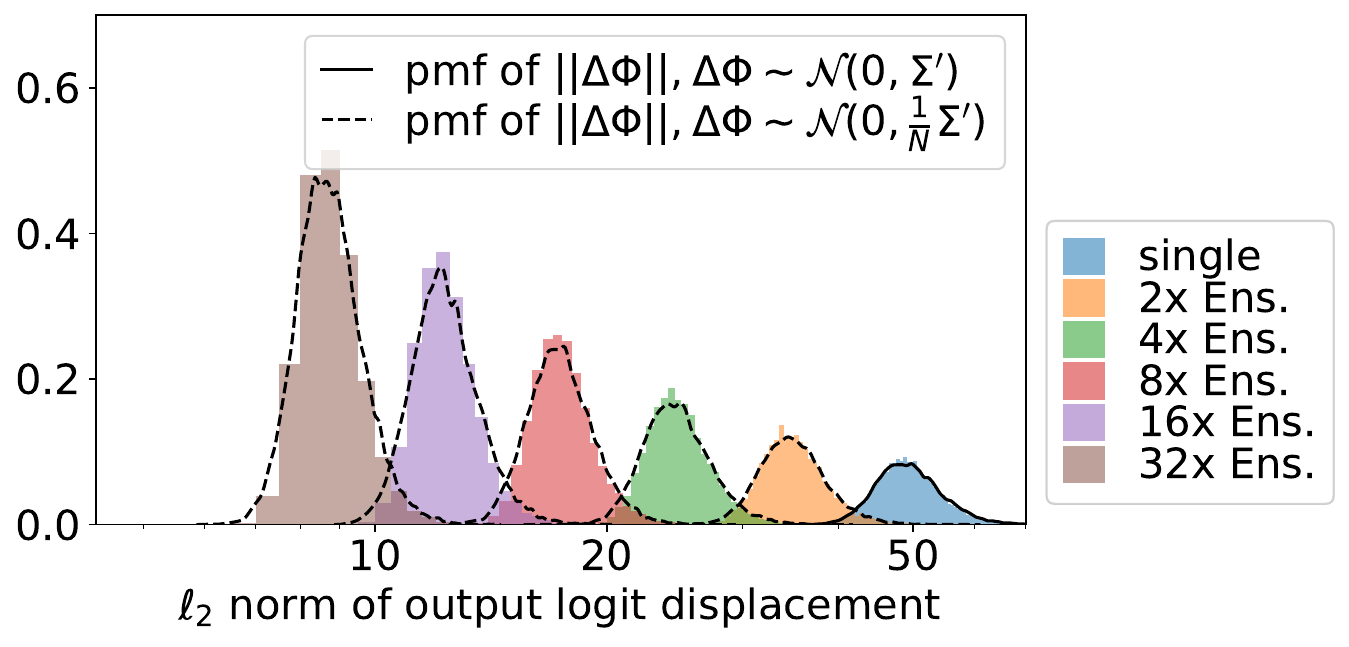}
			\label{supp_fig:hist_high_dim:logit_diff_337}
		}
		\subfloat[{\small \texttt{val\_00015098}.}]{
			\includegraphics[width=0.24\linewidth]{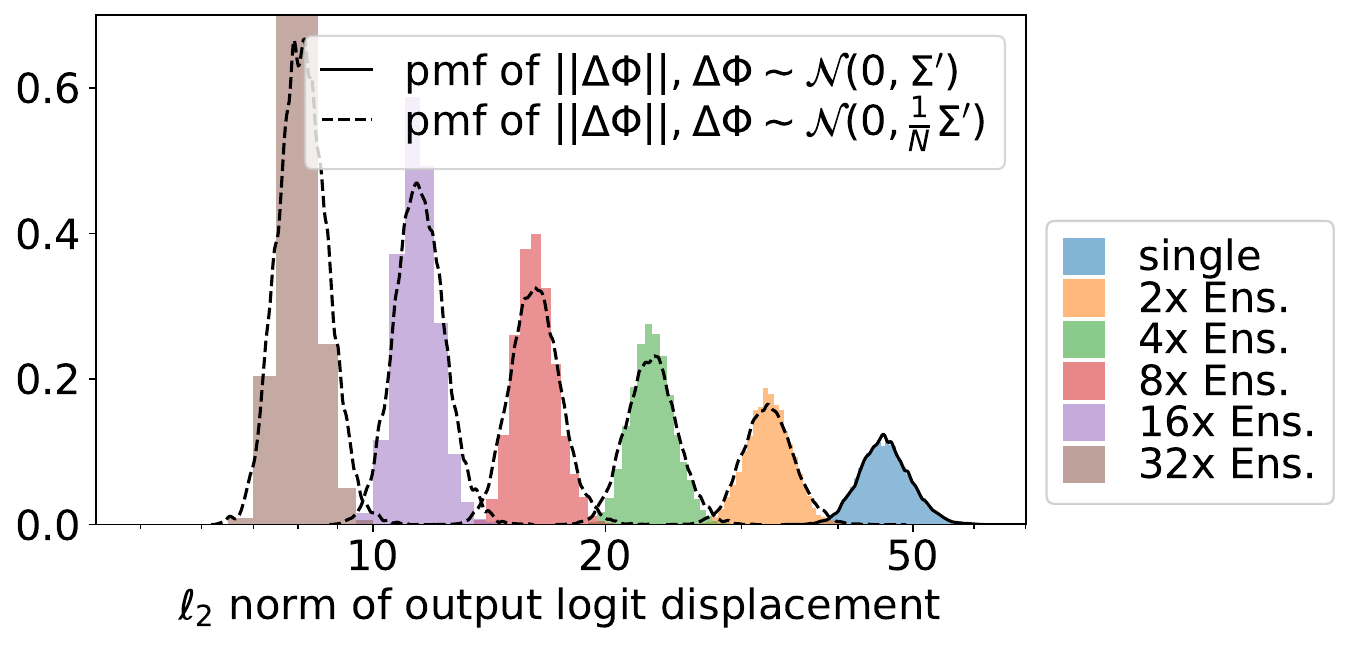}
			\label{supp_fig:hist_high_dim:logit_diff_399}
		}
		\subfloat[{\small \texttt{val\_00034619}.}]{
			\includegraphics[width=0.24\linewidth]{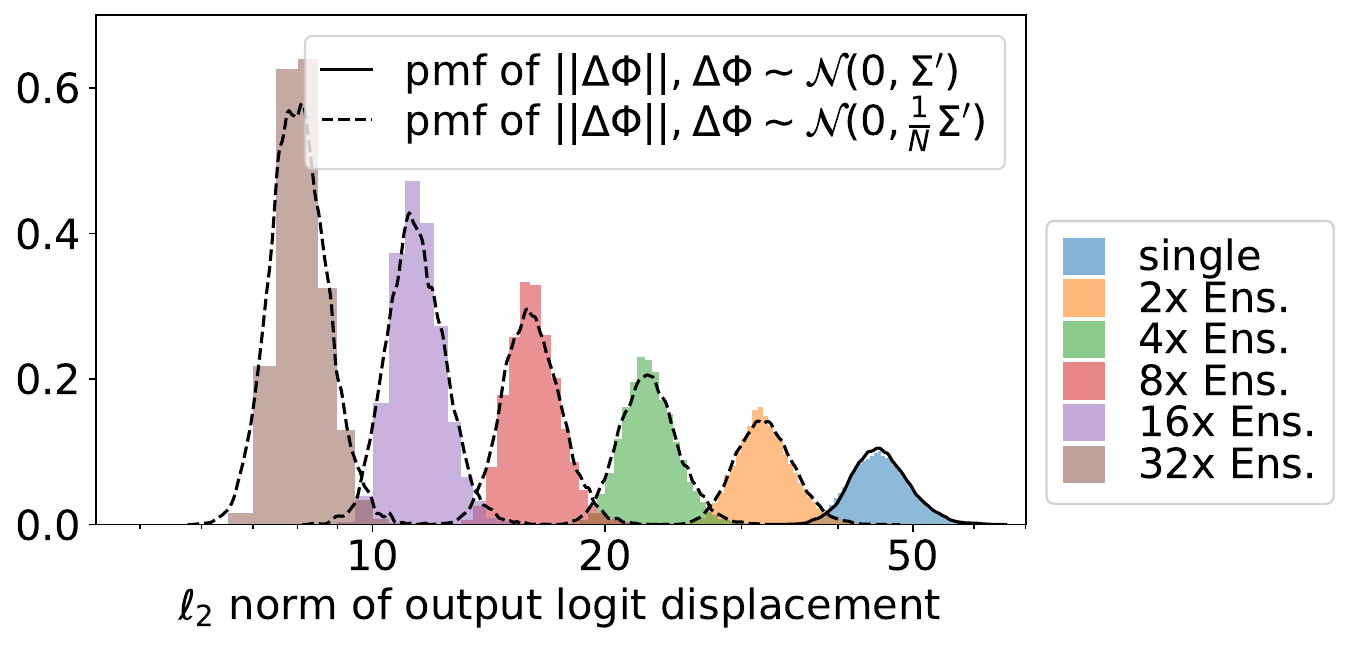}
			\label{supp_fig:hist_high_dim:logit_diff_217}
		}
		\subfloat[{\small \texttt{val\_00014560}.}]{
			\includegraphics[width=0.24\linewidth]{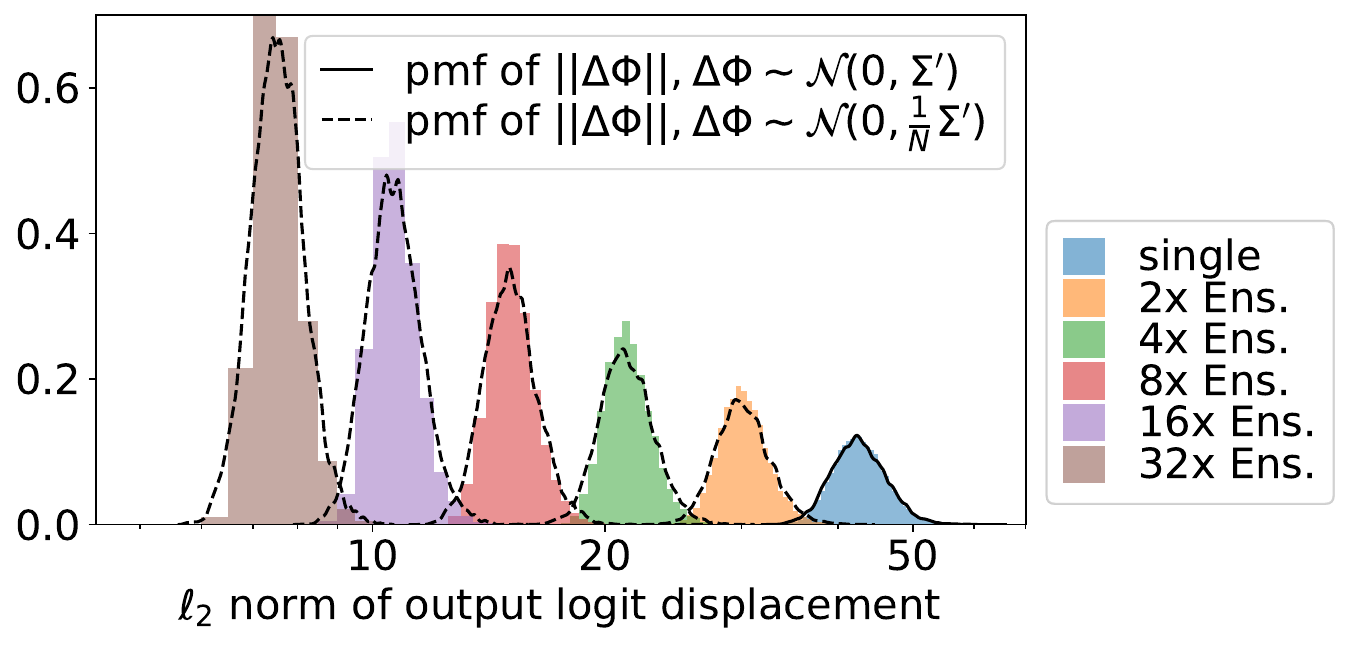}
			\label{supp_fig:hist_high_dim:logit_diff_431}
		}
		\caption{
				\textbf{$\ell_2$ norm histogram of logit displacement between two random ensembles.}
				The bin size is $0.5$.
				Two random ensembles are the same type (homogeneous ResNet-18 \versus homogeneous ResNet-18).
				$\Delta\vmu=\vmu_1-\vmu_2=\vzero, \mSigma'=2\mSigma_1=2\mSigma_2$.
		}
		\label{supp_fig:hist_high_dim:logit_diff}
	\end{minipage}
	\begin{minipage}{\textwidth}
		\centering
		\subfloat[{\small \texttt{val\_00009585}.}]{
			\includegraphics[width=0.24\linewidth]{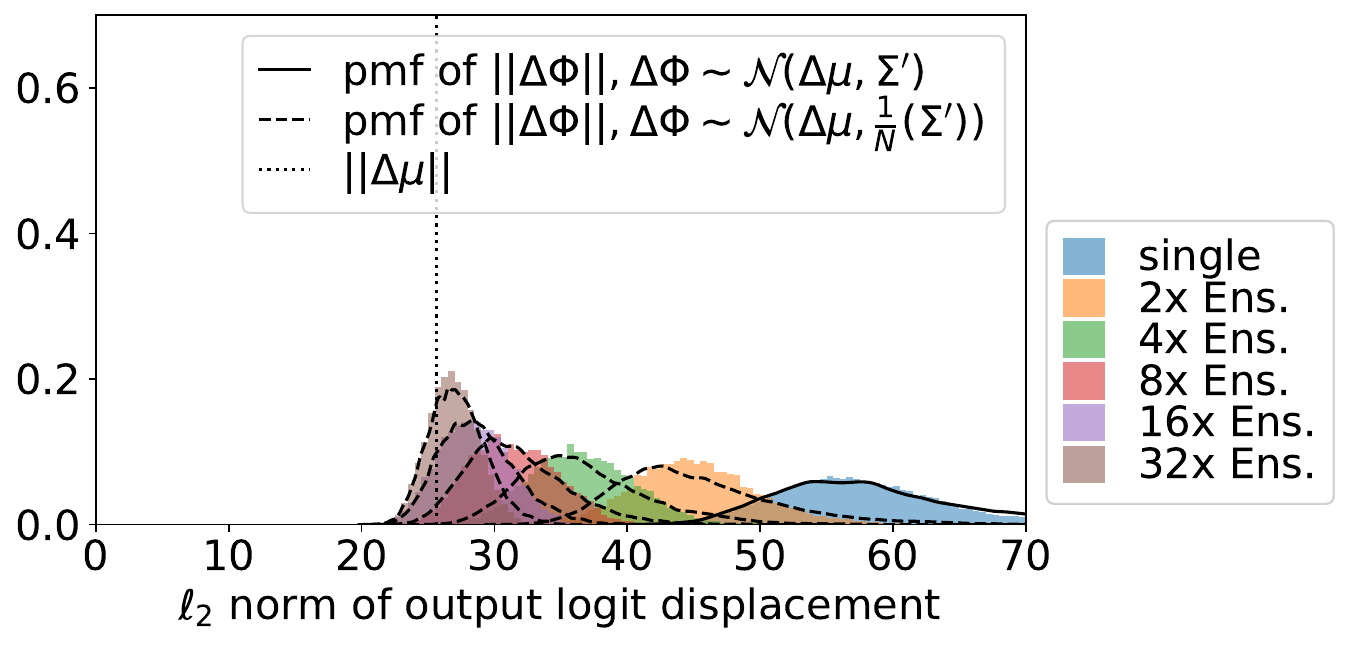}
			\label{supp_fig:hist_high_dim:logit_diff_hetero_337}
		}
		\subfloat[{\small \texttt{val\_00015098}.}]{
			\includegraphics[width=0.24\linewidth]{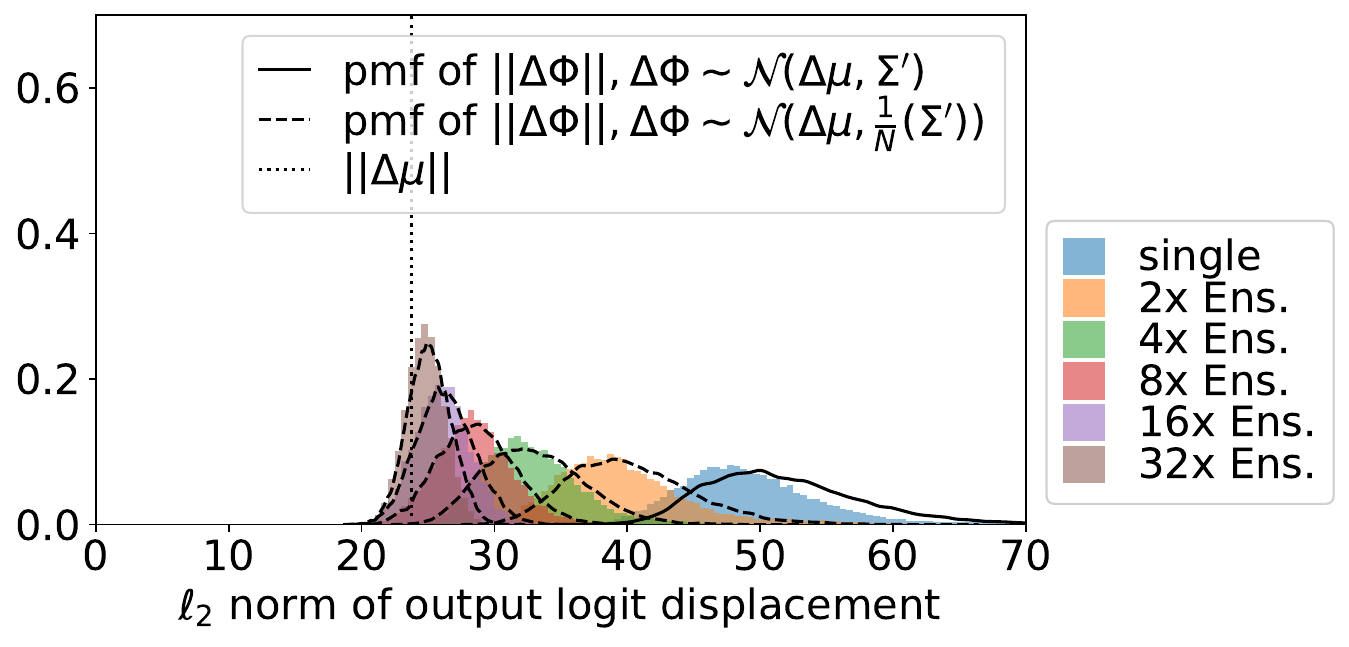}
			\label{supp_fig:hist_high_dim:logit_diff_hetero_399}
		}
		\subfloat[{\small \texttt{val\_00034619}.}]{
			\includegraphics[width=0.24\linewidth]{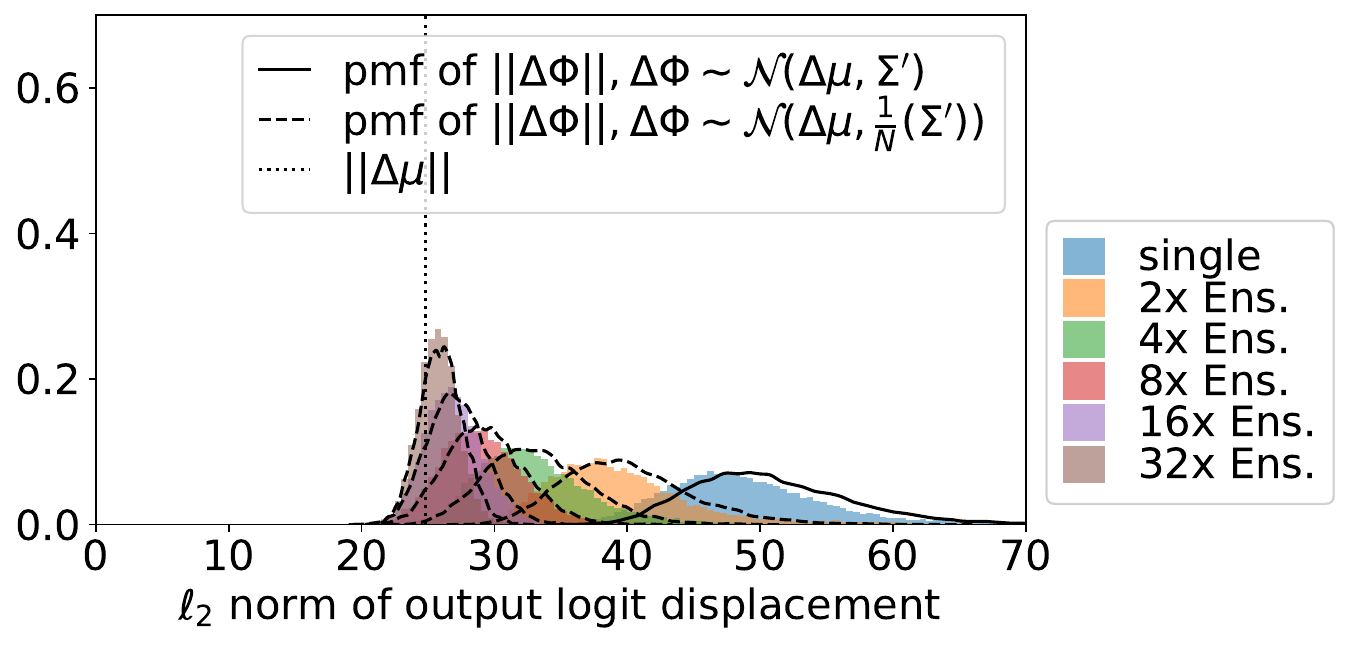}
			\label{supp_fig:hist_high_dim:logit_diff_hetero_217}
		}
		\subfloat[{\small \texttt{val\_00014560}.}]{
			\includegraphics[width=0.24\linewidth]{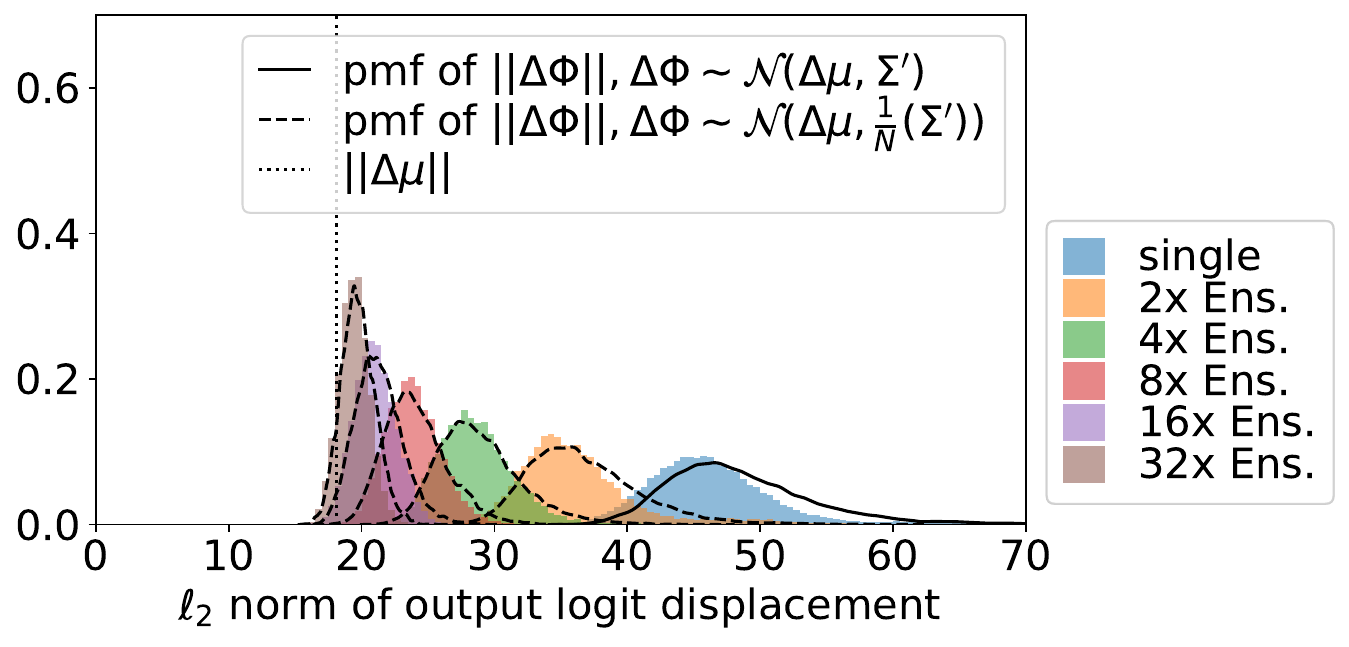}
			\label{supp_fig:hist_high_dim:logit_diff_hetero_431}
		}
		\caption{
				\textbf{$\ell_2$ norm histogram of logit displacement between two random ensembles.}
				The bin size is $0.5$. Two random ensembles are of different types (homogeneous ResNet-18 vs homogeneous ResNet-50).
				$\Delta\vmu=\vmu_1-\vmu_2\neq\vzero, \mSigma'=\mSigma_1+\mSigma_2$.
		}
		\label{supp_fig:hist_high_dim:logit_diff_hetero}
	\end{minipage}
\end{figure*}

To verify our hypothesis in higher dimension space, we train a standard ResNet-18 on the {\em full} ImageNet with 256 random seeds. We take the images in \cref{supp_fig:more_data_example} as input and illustrate the $\ell_2$ norm histogram of logit displacement between two random ensembles with different ensemble sizes in \cref{supp_fig:hist_high_dim:logit_diff}. For the cases where two ensembles have different architectures, we train a standard ResNet-50 on and observe the $\ell_2$ norm histogram of logit displacement between a random ResNet-18 ensemble and a random ResNet-50 ensemble with different ensemble sizes. The results are shown in \cref{supp_fig:hist_high_dim:logit_diff_hetero}.

\section{Representation Landscape of PC-Training at the Feature Level}
\label{appendix:feature}

We have discussed the representation landscape of PC-Training in~\cref{sec:probe:landscape} at the \emph{logit} space and provide some more data points above.
The analysis can be done in the \emph{feature} space as well.
The main challenge is that features from two arbitrary models are not directly comparable
and we address this \emph{feature interoperability} issue by Backward-Compatible Training~\cite{shen2020bct}.
We first introduce the BCT method and then derive that formulation to attain the feature at the \emph{penultimate layer} of an ensemble. 
Based on these we can analyze two-dimensional examples and the higher-dimension validation experiments. 

\myparagraph{Preliminaries.}
Shen~\etal propose an approach termed BCT~\cite{shen2020bct} to align two arbitrary deep models so that the embeddings are interoperable with each other.
Formally speaking, a model $\model$ includes an embedding module ($\vz=\cF(x)$,~\aka~\emph{backbone}) and a classification layer ($\vs=\cH(\vz)$,~\aka~\emph{head}) on top,
~\ie $\model(x) = \vphi(x) = (\cH\circ\cF)(x) $.
Given a reference model $\model^\mathrm{(ref)}$,
BCT imposes a loss term so that two model heads are close,~\ie
$\cH^\mathrm{(bct)}_i\sim\cH^\mathrm{(ref)}$\footnote{In fact if we assume that $\cH^\mathrm{(bct)}$ and $\cH^\mathrm{(ref)}$ have the same shape, we can also do as follows:
	we train $\model^\mathrm{(ref)}$ and then $\model^\mathrm{(bct)}$ with parameters randomly initialized except the head copies weights from $\cH^\mathrm{(ref)}$ and is \emph{fixed}.
	Nevertheless, we follow BCT's formulation since it is more generic.
}.
As a result, $\cF^\mathrm{(bct)}(x)$ and $ \cF^\mathrm{(ref)}(x)$ lie in the same vector space and are thus comparable,~\ie$\cF^\mathrm{(bct)}(x)\sim\cF^\mathrm{(ref)}(x)$, regardless of the underlying architecture.

\myparagraph{Ensemble of many feature-interoperable models.}
It is noteworthy that feature interoperability does \textbf{not} affect NFR as reported in~\cite{yan2021pct}.
We also re-validate that two models, $\cF^\mathrm{(bct)}_1(x)$ and  $\cF^\mathrm{(bct)}_2(x)$, trained using BCT \wrt $\model^\mathrm{(ref)}$ have similar NFR compared to two without BCT.
However, their features are comparable,~\ie$\cF^\mathrm{(bct)}_1(x)\sim\cF^\mathrm{(bct)}_2(x)\sim\cF^\mathrm{(ref)}(x)$.
So is any linear combination in between.

The arguments hold when the number of feature-interoperable models $n$ increases.
Therefore, if we write down their averaged logits, we can factor out the head,~\ie
{\small
	\begin{align}
		& \vphi^\mathrm{(bct,ens)}(x) 
		= \frac{1}{N} \sum_n \vphi^\mathrm{(bct)}_n(x) =\frac{1}{N} \sum_n \left(\cH^\mathrm{(bct)}_n\circ\cF^\mathrm{(bct)}_n\right)(x) \\
		&\approx \frac{1}{N} \sum_n \left(\cH^\mathrm{(ref)}\circ\cF^\mathrm{(bct)}_n\right)(x) 
		= \cH^\mathrm{(ref)}\circ \left( \frac{1}{N} \sum_n\cF^\mathrm{(bct)}_n(x) \right).
	\end{align}
}
It implies that the averaged feature can be viewed as this ensemble's feature,~\ie $\cF^\mathrm{(bct,ens)}(x)=\frac{1}{N}\sum \cF^\mathrm{(bct)}_n(x)$.

\begin{figure*}[t]
	\begin{center}
		\subfloat[{\small Two single models: Test samples flip even if not close to the boundary ({\color{magenta}long arrows}).}]{
			\includegraphics[width=0.22\linewidth]{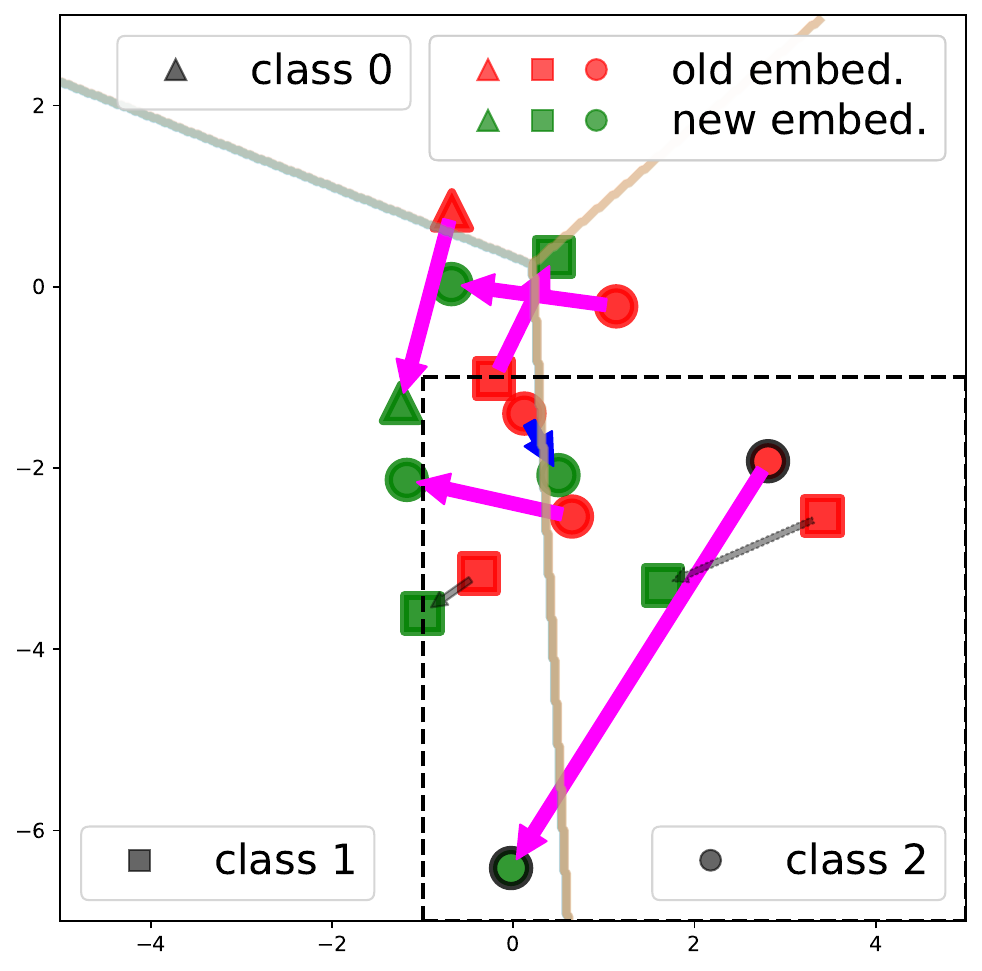}
			\label{supp_fig:toy_model_single}
		}
		\hskip 0.08 in
		\subfloat[{\small Two ensembles of 3 models each (3$\times$): Few samples that are far from the boundary flip ({\color{magenta}shorter arrows}).}]{
			\includegraphics[width=0.22\linewidth]{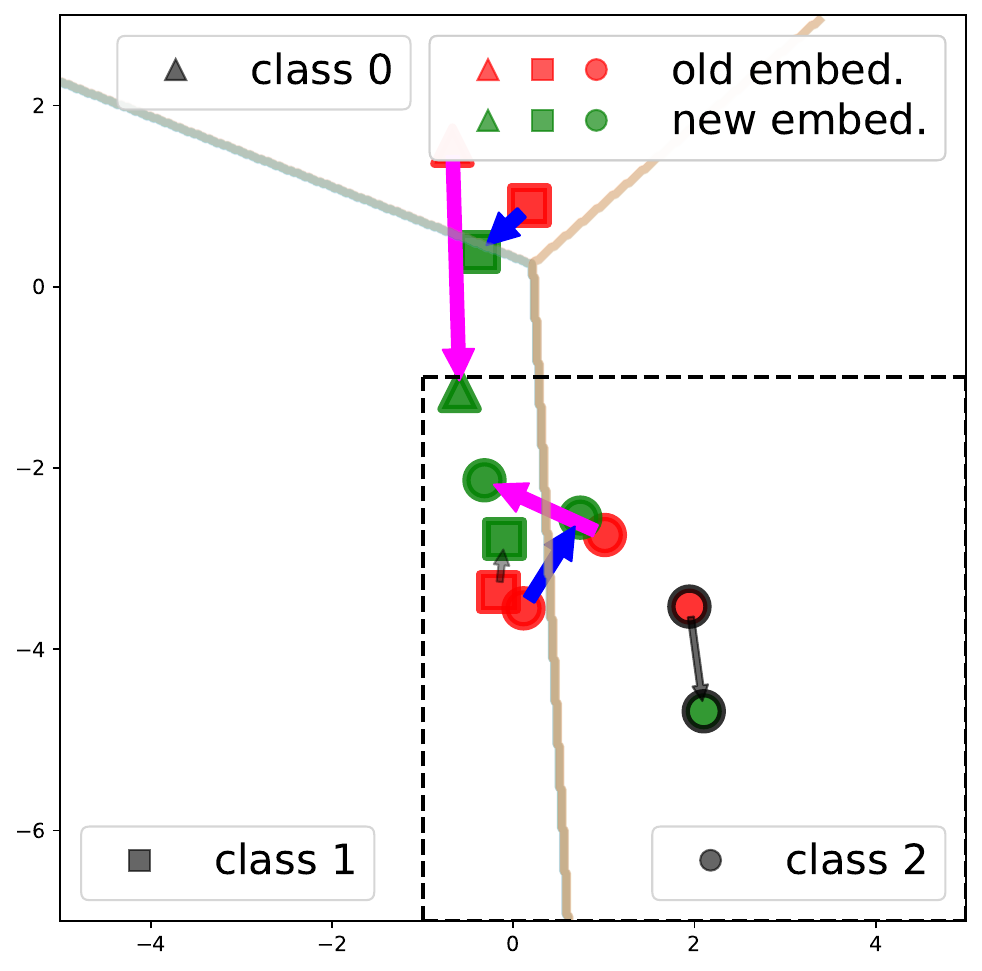}
			\label{supp_fig:toy_model_ensemble}
		}
		\hskip 0.08 in
		\subfloat[{\small Embeddings of ensembles and their members: Individual models' embeddings (lighter circles) center around the mode.}]{
			\includegraphics[width=0.22\linewidth]{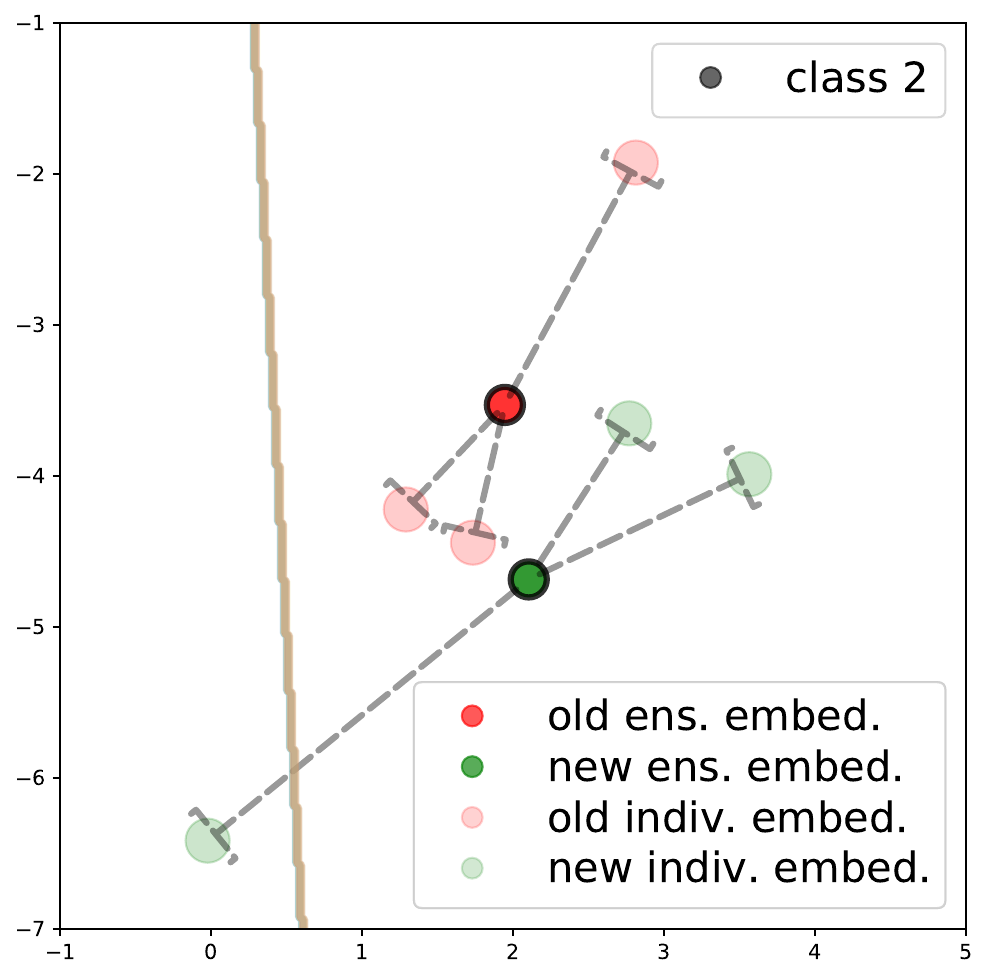}
			\label{supp_fig:toy_model_drift}
		}
		\hskip 0.08 in
		\subfloat[{\small PMF of feature difference.}]{
			\includegraphics[width=0.21\linewidth]{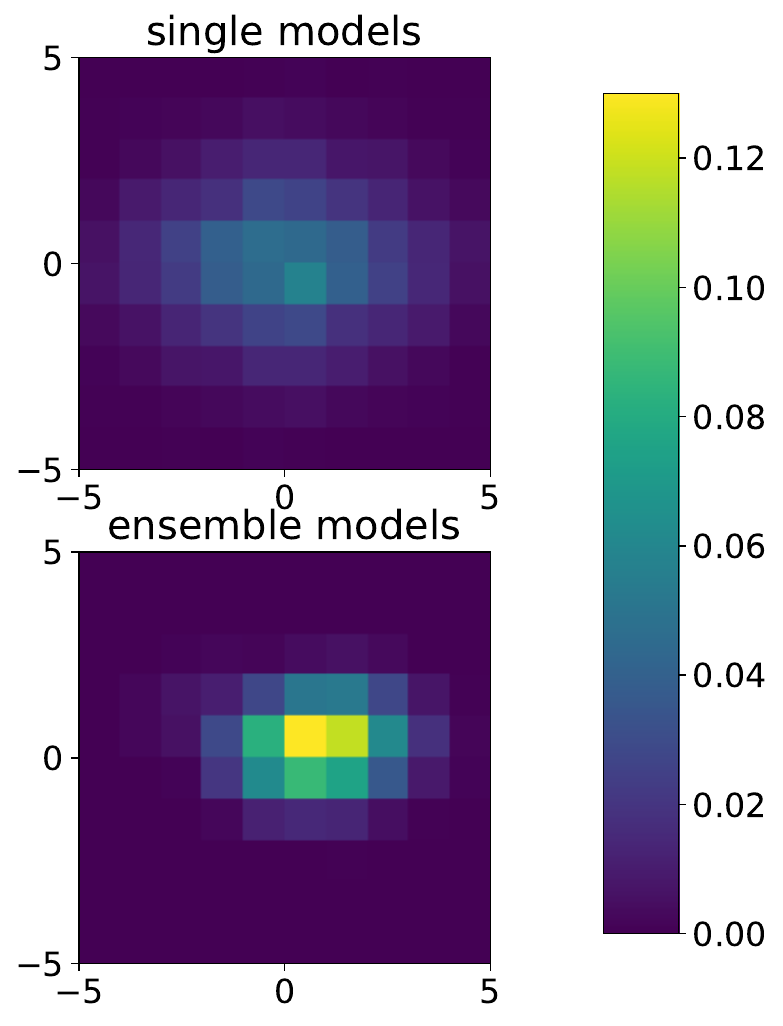}
			\label{supp_fig:hist2d_clt}
		}
		\caption{
				\textbf{Visualization of a 3-class 2-dimensional example}.
				\textbf{(a-c)}: 2D feature embedding of two single models or ensembles. 
				\trimark, \smark, and  $\bullet$ refer to the ground-truth classes for each sample.
				{\textcolor{tabred}{Red}} and {\textcolor{tabgreen}{green}} data points refer to old and new model's embeddings.
				{\textcolor{magenta}{Magenta arrow}}, {\textcolor{blue}{blue arrow}}, {\textcolor{Darkgray}{gray arrow}}
				link negative flip, positive flip, and consistent (either both correct or both wrong) prediction pairs.
				All dots with black borders depict the same image.
				\textbf{(d)}: Estimated probability mass function (PMF) of feature difference between two single models or ensembles $\Delta\vz$.
				The $x$- and $y$-axes denote the 2D feature difference.
				The heatmap value denotes the estimated probability density.
				The ensemble's co-variance is significantly smaller than the single model.
				The figure is best viewed in color.}
		\label{supp_fig:toy_example}
	\end{center}
    \vskip -0.1 in
\end{figure*}

\begin{figure*}[t]
	\begin{center}
		\centering
		\subfloat[{\small \texttt{val\_00009585}.}]{
			\includegraphics[width=0.24\linewidth]{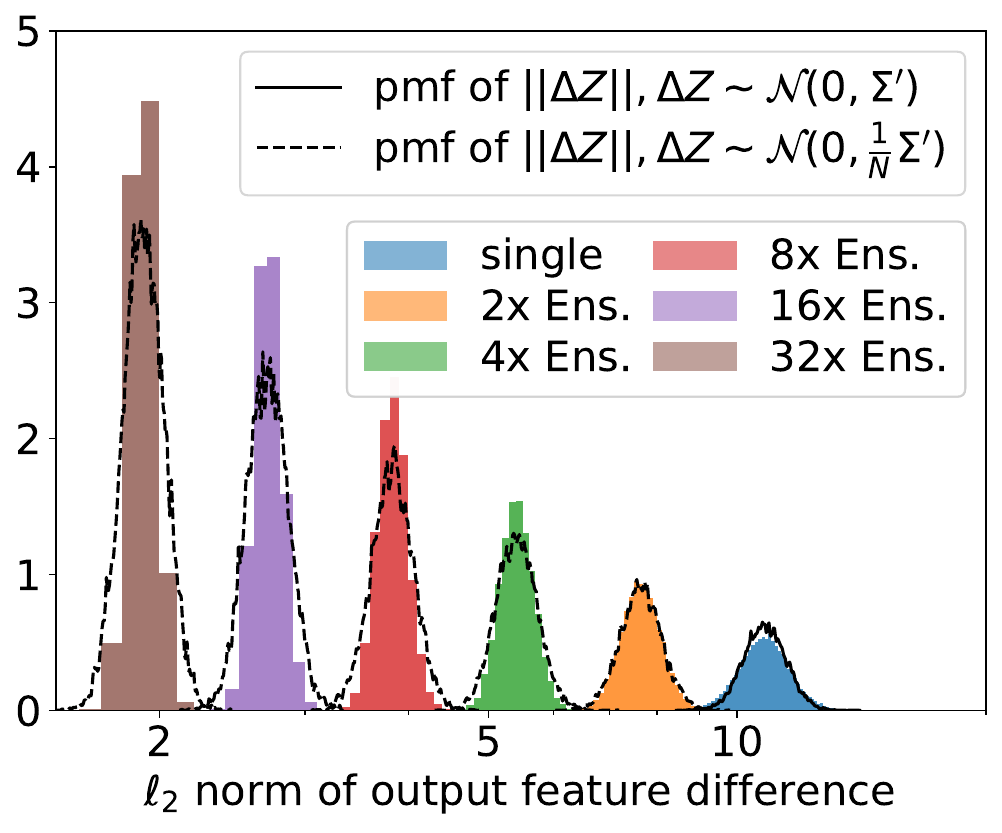}
			\label{fig:hist_high_dim:feat_diff_337}
		}
		\subfloat[{\small \texttt{val\_00015098}.}]{
			\includegraphics[width=0.24\linewidth]{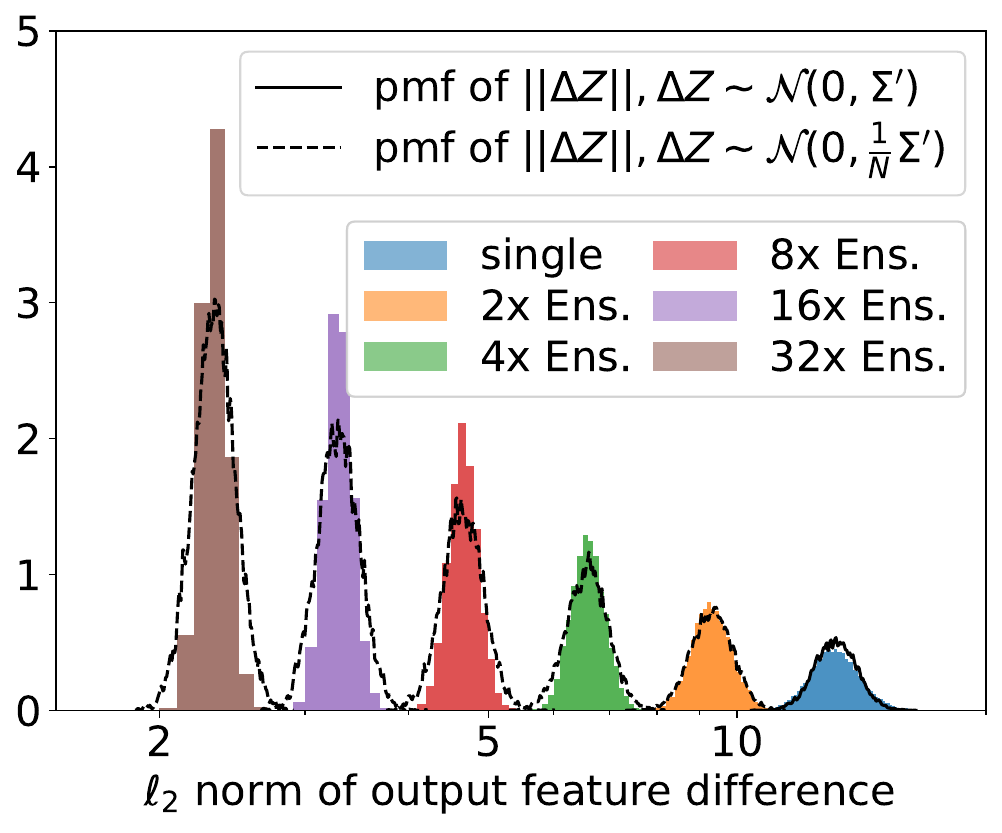}
			\label{fig:hist_high_dim:feat_diff_399}
		}
		\subfloat[{\small \texttt{val\_00034619}.}]{
			\includegraphics[width=0.24\linewidth]{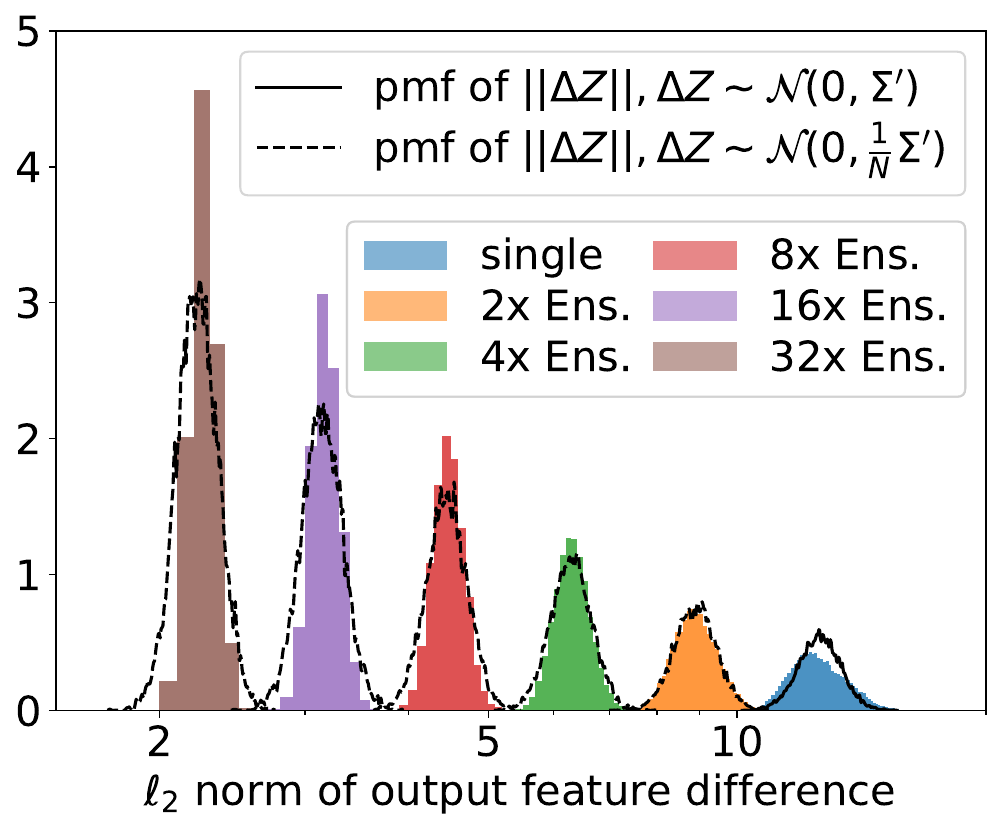}
			\label{fig:hist_high_dim:feat_diff_217}
		}
		\subfloat[{\small \texttt{val\_00014560}.}]{
			\includegraphics[width=0.24\linewidth]{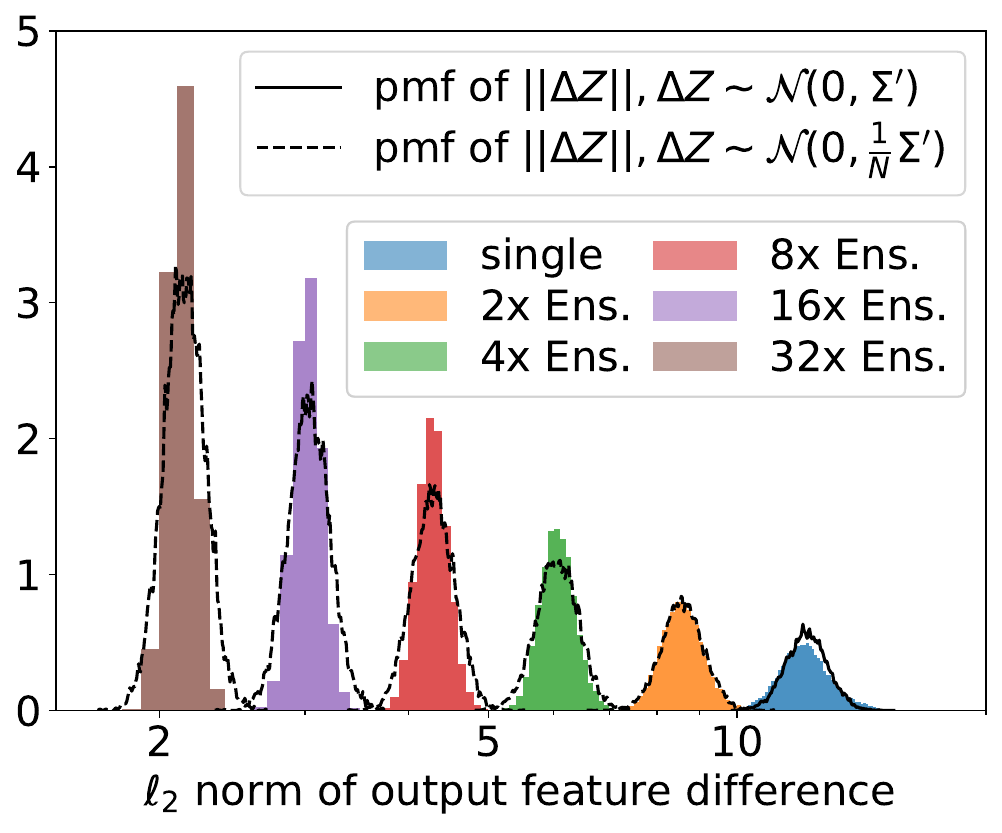}
			\label{fig:hist_high_dim:feat_diff_431}
		}
		\caption{
			\textbf{$\ell_2$ norm histogram of feature difference between two random ensembles.}
			Note that the bin size is $0.1$.
			Two random ensembles are of the same type (ResNet-18 \versus ResNet-18).
			$\Delta\vmu=\vmu_1-\vmu_2\neq\vzero, \mSigma'=\mSigma_1+\mSigma_2$.
			We also plot the simulated probability mass function (PMF):
			the solid line for the norm of a simulated normal distribution $\mathcal{N}\left(\Delta\vmu, (\mSigma_1+\mSigma_2)\right)$ whose parameters are estimated from all available single models;
			the dashed lines for extrapolated distribution $\mathcal{N}\left(\Delta\vmu, \frac{1}{m}(\mSigma_1+\mSigma_2)\right)$.
			Consistency between the ensembles' histograms and PMFs supports our hypotheses in~\cref{appendix:feature}.
		}
		\label{supp_fig:hist_high_dim:feat_diff}
	\end{center}
    \vskip -0.1 in
\end{figure*}

\myparagraph{A two-dimensional example.}
Similar to ~\cref{sec:prob:toy_example}, to illustrate the behavior of negative flips in the feature space, we create a toy example by selecting three classes\footnote{``Labrador retriever'' (n02099712), ``Weimaraner'' (n02092339), and ``French bulldog'' (n02108915).} from ImageNet~\cite{deng2009imagenet} and training a ResNet-18-like model with a slight modification such that the penultimate layer's dimension is changed to 2.
The feature level visualization is presented in \cref{supp_fig:toy_model_single} and~\ref{supp_fig:hist2d_clt}. We can observe similar observations as in the logit space after the penultimate layer features are aligned with BCT~\cite{shen2020bct}.

\myparagraph{Validations on higher dimensions.}
We repeat the high-dimensional validation in text on the penultimate layer features, the results are shown in~\cref{supp_fig:hist_high_dim:feat_diff}.
We see that the PMF curve fits the histogram of single models well, implying that features of these models could indeed follow a Normal distribution.
We conducted the same experiments above on many more images and the conclusion holds well.
If we move to ensembles of $m$ models each, the feature difference follows another normal distribution whose co-variance matrix is scaled by a factor of $m$,
~\ie $\Delta\vz^\mathrm{(ens)}\sim\mathcal{N}\left(\vzero,\frac{2}{m}\mSigma\right)$.
We demonstrate that the rest of the histograms are indeed consistent with the estimated PMF of $\| \Delta\vz^\mathrm{(ens)} \|^2$ (dashed lines in \cref{supp_fig:hist_high_dim:feat_diff}).

\end{document}